\documentclass{article}

\usepackage{fullpage}
\usepackage{latexsym,mathrsfs}
\usepackage{amssymb,amsfonts,amsmath,bm}
\usepackage{natbib}
\usepackage{algorithm}
\usepackage{algpseudocode}
\usepackage{algorithmicx}
\usepackage{color}
\usepackage{graphics}
\usepackage{graphicx}
\usepackage{bbm}

\def\ds{\displaystyle}
\def\R{\mathbb{R}}
\newcommand{\x}{\mathbf{x}}
\newcommand{\X}{\mathbf{X}}
\newcommand{\y}{\mathbf{y}}
\newcommand{\f}{\mathbf{f}}
\newcommand{\Y}{\mathbf{Y}}
\newcommand{\F}{\mathbf{F}}
\newcommand{\z}{\mathbf{z}}
\newcommand{\s}{\mathbf{x}}
\newcommand{\Sset}{\mathbb{X}}
\newcommand{\Rset}{\mathbb{R}}
\newcommand{\Xset}{\mathbb{X}}
\newcommand{\Prob}{\mathbb{P}}
\newcommand{\yr}{\mathcal{Y}}
\newcommand{\fr}{\mathcal{F}}
\newcommand{\Fr}{\boldsymbol{\mathcal{F}}}
\newcommand{\Yr}{\boldsymbol{\mathcal{Y}}}

\newcommand{\sr}{\mathbb{X}}

\newcommand{\cand}{_{\text{cand}}}
\newcommand{\esp}{\mathbb{E}}

\def\RR{\textsf{R}\/}
\def\cov{\operatorname{cov}}
\def\diag{\operatorname{diag}}

\newtheorem{definition}{Definition}
\newtheorem{theorem}{Theorem}

\begin{document}

\title{A Bayesian optimization approach to find Nash equilibria}

\author{Victor Picheny\footnote{MIAT, Universit\'e de Toulouse, INRA, Castanet-Tolosan, France (victor.picheny@inra.fr)}        \and
        Mickael Binois\footnote{The University of Chicago Booth School of Business, Chicago IL (mickael.binois@chicagobooth.edu)}  \and
        Abderrahmane Habbal\footnote{Universit\'e C\^ote d'Azur, Inria, CNRS, LJAD, UMR 7351, Nice, France (habbal@unice.fr)}
}

\maketitle

\begin{abstract}
Game theory finds nowadays a broad range of applications in engineering and machine learning.
  However, in a derivative-free, expensive black-box context, very few algorithmic solutions are 
  available to find game equilibria. Here, we propose a novel Gaussian-process based approach for solving games
  in this context. We follow a classical Bayesian optimization framework, with sequential 
  sampling decisions based on acquisition functions. Two strategies are proposed, 
  based either on the probability of achieving equilibrium or on the Stepwise Uncertainty Reduction paradigm. 
  Practical and numerical aspects are discussed in order to enhance the scalability and reduce computation time.
  Our approach is evaluated on several synthetic game problems with varying number of players and decision space dimensions.
  We show that equilibria can be found reliably for a fraction of the cost (in terms of 
  black-box evaluations) compared to classical, derivative-based algorithms.
  The method is available in the \RR{} package \texttt{GPGame} available on CRAN at 
\texttt{https://cran.r-project.org/package=GPGame}.

\textbf{Keywords: } Game theory, Gaussian processes,  Stepwise Uncertainty Reduction
\end{abstract}

\section{Introduction}

Game theory arose from the need to model economic behavior, where multiple decision makers (MDM) with antagonistic goals is a natural feature.
It was further extended to broader areas, where MDM had however to deal with systems governed by ordinary differential equations, the so-called differential games. 
See e.g., \citet{Gibbons-Game-92} for a nice introduction to the general theory and \citet{MR0210469} for differential games.
Recently, engineering problems with antagonistic design goals and with real or virtual MDM were formulated 
by some authors within a game-theoretic framework.  
See e.g., \citet{JAD2014-HELICO} for aerodynamics, \citet{Habbal2004-A1} for structural topology design, 
\citet{HabbalSICON2013} for missing data recovery problems. 
The study of multi-agent systems or games such as poker under this setting is also quite common in the AI and machine learning communities, see e.g., 
\citet{Johanson2009,Lanctot2012,Brown2015}.

Solutions to games are called equilibria. Contrarily to classical optimization, the definition of an equilibrium depends on the game setting (or rules).
Within the static with complete information setting, a relevant one is the so-called Nash equilibrium (NE). 
Shortly speaking, a NE is a fixed-point of iterated many single optimizations 
(see the exact definition in Section-\ref{sec:background} below). 
Its computation generically carries the well known tricks and pitfalls related to computing a fixed-point, 
as well as those related to intensive optimizations notably when cost evaluations are expensive, 
which is the case for most engineering applications. There is an extensive literature related to theoretical analysis of algorithms for computing NE \citep{MR942837,MR899829,Rubinstein}, 
but very little -if any- on black-box models (i.e., non convex utilities) and expensive-to-evaluate ones; to the best of our knowledge, only home-tailored implementations are used. 
On the other hand, \textit{Bayesian optimization} \citep[BO,][]{mockus1989} is a popular approach to tackle black-box problems. Our aim is to investigate the extension of 
such approach to the problem of computing game equilibria. 

BO relies on Gaussian processes, which are used as emulators (or surrogates) of the black-box model outputs based on a small set of 
model evaluations. Posterior distributions provided by the Gaussian process are used to design \textit{acquisition functions}
that guide sequential search strategies that balance between exploration and exploitation.
Such approaches have been applied for instance to multi-objective problems \citep{wagner2010expected}, as well as 
transposed to frameworks other than optimization, such as uncertainty quantification \citep{bect2012sequential}
or optimal stopping problems in finance \citep{Gramacy2015}.

In this paper, we show that the BO apparatus can be applied to the search of game equilibria,
and in particular the classical Nash equilibrium (NE). To this end, 
 we propose two complementary acquisition functions, one based on a greedy search approach and one based on the 
Stepwise Uncertainty Reduction paradigm \citep{fleuret1999graded}.
The corresponding algorithms require very few model evaluations to converge to the solution.
Our proposal hence broadens the scope of applicability of equilibrium-based methods, 
as it is designed to tackle derivative-free, non-convex and expensive models, for which a game perspective
was previously out of reach.

The rest of the paper is organized as follows. Section \ref{sec:background} reviews the basics of 
game theory and presents our Gaussian process framework. 
Section \ref{sec:sequential} presents our main contribution, with the definition of 
two acquisition functions, along with computational aspects. 
Finally, Section \ref{sec:experiments} demonstrates the capability of our algorithm
on three challenging problems.

\section{Background}\label{sec:background}
\subsection{Games and equilibria}
\subsubsection{Nash games}
We consider primarily the standard (static, under complete information) Nash equilibrium problem \citep[NEP,][]{Gibbons-Game-92}. 
\begin{definition}
 A NEP consists of $p \ge 2$ decision makers (i.e., players), 
where each player $i \in \{1, \ldots, p\}$ tries to solve his optimization problem:
\begin{equation}\label{eq:NEP}
 (\mathcal{P}_i) \quad \min_{\s_i \in \sr_i} y_i(\s),
\end{equation}
where $\y(\x) = \left[y_1(\x),\ldots, y_p(\x)\right]: \sr \subset \Rset^n \rightarrow \Rset^p$ (with $n \geq p$) denotes a vector of cost functions
(a.k.a. pay-off or utility functions), $y_i$
denotes the specific cost function of player $i$, 
and the vector $\s$ consists of block components $\s_1, \ldots, \s_p$ $\left(\s = (\s_j)_{1\le j \le p}\right)$. 
\end{definition}
Each block $\s_i$ denotes the variables of player $i$ and $\sr_i$ its corresponding action space and $\sr = \prod_i \sr_i$.
We shall use the convention $y_i(\s) = y_i(\s_i, \s_{-i})$ when we need to emphasize the role of $\s_i$. 

\begin{definition}
 A Nash equilibrium $\s^* \in \sr$ is a strategy such that:
\begin{equation} \label{eq:NEdef}
(NE) \quad \forall i,\ 1\le i \le p, \quad  \s^*_i  \in \arg \min_{\s_i \in \sr_i } y_i(\s_i, \s^*_{-i}).
\end{equation}
\end{definition}
In other words, when all players have chosen to play a NE, then no single player has incentive to move from his $\s^*_i$.
Let us however mention by now that, generically, Nash equilibria are not efficient, i.e., do not belong to the underlying set of best compromise solutions, 
called Pareto front, of the objective vector $(y_i(\s))_{\s\in \sr}$.

\subsubsection{Random games}
We shall also deal with the case where cost functions are uncertain. Such problems belong to a family of random games that are called disturbed games by Harsanyi in \citep{harsanyi1973games}.
We denote such cost functions $f_i(\x, \boldsymbol{\epsilon}(\xi))$, where $\boldsymbol{\epsilon}=( \epsilon_i): \Xi \rightarrow \Rset^p$
is a random vector defined over a probability space $ (\Xi,\mathcal{F}, \Prob)$. 
In the following we refer to our setting as random games, but emphasize that we consider {\it static} Nash games with expectations of randomly perturbed costs.

\begin{definition}
  Assuming risk-neutrality of the players, a random Nash game consists of $p \ge 2$ players, where each player $i \in \{1, \ldots, p\}$ tries to solve
\begin{equation}\label{eq:SNEP}
 (\mathcal{SP}_i) \quad \min_{\s_i \in \sr_i} \esp[f_i(\s, \boldsymbol{\epsilon}(\xi))].
\end{equation}
\end{definition}

\begin{definition}\label{SNE}
  A random Nash equilibrium $\s^* \in \sr$ is a strategy such that:
\begin{equation} \label{eq:SNEdef}
(SNE) \quad \forall i,\ 1\le i \le p, \quad  \s^*_i  \in \arg \min_{\s_i \in \sr_i } \esp[f_i(\s_i, \s^*_{-i}, \boldsymbol{\epsilon}(\xi))].
\end{equation}
\end{definition}
Note that setting $y_i = \esp[f_i(\s, \boldsymbol{\epsilon}(\xi))]$, we see directly that $(NE)$ and $(SNE)$ are equivalent.

\subsubsection{Working hypotheses}
In this work, we focus on continuous-strategy non-cooperative Nash games (i.e., with infinite sets $\sr_i$) or on large finite games (i.e., with large finite sets $\sr_i$). \\

Our working hypotheses are: 
\begin{itemize}
 \item queries on the cost function (i.e., pointwise evaluation of the $y_i$'s for a given $\s$)
result from an expensive process: typically, the $y_i$'s can be the outputs of numerical models;
 \item the cost functions may have some regularity properties but are possibly strongly not convex (e.g., continuous and multimodal);
 \item the cost functions evaluations can be corrupted by noise;
 \item $\Xset$ is either originally discrete, or a representative discretization of it is available (so that the equilibrium of the corresponding finite game is similar to the one of original problem). 
\end{itemize}
Note that to account for the particular form of NEPs, $\Xset$ must realize a full-factorial design: $\sr=\sr_1\times \ldots \times \sr_p$.
Given each action space $\Sset_i=\{\s_i^1, \ldots, \s_i^{m_i} \}$ 
of size $m_i$,
$\Xset$ consists of all the combinations $(\s_i^k, \s_j^l)$ 
($1 \leq i \neq j \leq p$, $1 \leq k \leq m_i$, $1 \leq l \leq m_j$),
and we have $N := \operatorname{Card}(\Xset) = \prod_{i=1}^p m_i$.

Let us remark that we do not require for an equilibrium to exist and be unique, the case when there is no or equilibria is discussed in Section \ref{ssec:conver}.

In the case of noisy evaluations, we consider only here an additive noise corruption:
\begin{equation}
  f_i(\x,\boldsymbol{\epsilon}(\xi)) = y_i(\x) + \epsilon_i(\xi),
\end{equation}
and we assume further that $\boldsymbol{\epsilon}$ has independent Gaussian centered elements: $\epsilon_i \sim \mathcal{N}(0, \tau_i^2)$.
Notice that in this case, both problems and equilibria coincide, and can be solved with the same algorithm. 
Hence, in the following, all calculations are given in the noisy case, 
while the deterministic case of the standard NEP is recovered by setting $ \epsilon_i = 0$ and $\tau_i = 0$. 

Furthermore, we consider solely pure-strategy Nash equilibria \citep[as opposed to mixed-strategies equilibria, 
in which the optimal strategies can be chosen stochastically, see][Chapter 1]{Gibbons-Game-92}, and as such, 
we avoid solving the linear programs LP or linear complementarity problems LCP generally used in the dedicated classes of algorithms 
\`a la Lemke-Howson \citep{rosenmuller1971generalization}.

\subsubsection{Related work}
Let us mention that in the continuous (in the sense of smooth) games setting, there is an extensive literature dedicated to the computation of NE, based on the 
rich theory of variational analysis, starting with  the classical fixed-point algorithms to solve NEPs \citep{Rubinstein,MR942837,MR899829}. When the players share common constraints, Nash equilibria are shown to be Fritz-John (FJ) points \citep{dorsch2013structure}, which allows the \emph{op. cit.} authors to propose a nonsmooth projection method (NPM) well adapted for the computation of FJ points. From other part, it is well known (and straightforward) that Nash equilibria are in general not classical Karush-Kuhn-Tucker (KKT) points, nevertheless, a notion of KKT condition for generalized Nash equilibria GNEP is developed in \citet{kanzow2016augmented}, which allows the authors to derive an augmented Lagrangian method to compute GNEPs.

Noncooperative stochastic games theory, starting from the seminal paper by Shapley \citep{shapley1953stochastic}, occupies nowadays most of the game theorists, and a vast  literature is dedicated to stochastic differential games \citep{friedman1972stochastic}, robust games \citep{nishimura2009robust}, games on random graphs,  or agents learning games \citep{hu2003nash}, among many other branches, and it is definitely out of the scope of the paper to review all aspects of the field. 
See also the introductory book \citet{neyman2003stochastic} to the basic -yet deep- concepts of the stochastic games theory.

We also do not consider games with additional specific structures, like cooperative, zero-sum stochastic or deterministic games, or repeated  Nash games \citep{littman2005polynomial}. For these games, tailored algorithms should be used, 
among which are the pure exploration statistical learning with Monte Carlo Tree Search \citep{garivier2016maximin} or multi-agent reinforcement learning MARL, see e.g., \citet{games2016lenient} and references therein.

We stress here that none of the above-mentioned approaches are designed to tackle expensive black-box problems. 
They may even prove unusable in this context, either because they could simply not converge or require too many cost function evaluations to do so.

\subsection{Bayesian optimization}
\subsubsection{Gaussian process regression}
The idea of replacing an expensive function by a cheap-to-evaluate surrogate is not recent,
with initial attempts based on linear regression. 
Gaussian process (GP) regression, or kriging, extends the versatility and efficiency of surrogate-based methods in many applications,
such as in optimization or reliability analysis. 
Among alternative non-parametric models such as radial basis functions or random forests, see e.g., \citet{Wang2007,Shahriari2016} for a discussion,
GPs are attractive in particular for their tractability, 
since they are simply characterized by their mean $m$ and covariance (or kernel) $k$ functions, see e.g., \citet{Cressie1993,Rasmussen2006}. 
In the following, we consider zero-mean processes ($m = 0$) for the sake of conciseness.

Briefly, for a single objective $y$, conditionally on $n$ noisy observations $\f = (f_1, \ldots, f_n)$, with independent, centered, Gaussian noise,
that is, $f_i = y(\x_i) + \varepsilon_i$ with $\varepsilon_i \sim \mathcal{N}(0, \tau_i^2)$,
the predictive distribution of $y$ is another GP, with mean and covariance functions given by:
\begin{eqnarray}
\mu(\x) & =& \textbf{k}(\x)^\top \textbf{K}^{-1} \f,  \\
\sigma^2(\x, \x') & =& k(\x, \x') - \textbf{k}(\x)^\top \textbf{K}^{-1} \textbf{k}(\x'),
\end{eqnarray}
where $\textbf{k}(\x) := (k(\x, \x_1), \dots, k(\x, \x_n))^\top$ and $\textbf{K} := (k(\x_i, \x_j) + \tau_i^2 \delta_{i=j})_{1 \leq i,j \leq n}$,
$\delta$ standing for the Kronecker function.
Commonly, $k$ belongs to a parametric family of covariance functions such as the Gaussian and Mat\'ern kernels, 
based on hypotheses about the smoothness of $y$. Corresponding hyperparameters are often obtained as maximum likelihood estimates, 
see e.g., \citet{Rasmussen2006} or \citet{Roustant2012} for the corresponding details.    

With several objectives, a statistical emulator for $\y(\x) = \left[y_1(\x),\ldots, y_p(\x)\right]$ is needed.
While a joint modeling is possible \citep[see e.g.,][]{Alvarez2011}, it is more common practice to treat the $y_i$'s separately. 
Hence, conditioned on a set of vectorial observations $\{\f_1, \ldots, \f_n\}$, 
our emulator is a multivariate Gaussian process $\Y$:
\begin{eqnarray}\label{eq:Y}
 \Y(.) \sim \mathcal{GP} \left( \boldsymbol{\mu}(.), \boldsymbol{\Sigma}\left(.,.\right) \right),
\end{eqnarray}
with $\boldsymbol{\mu}(.) = \left[\mu_1(.), \dots, \mu_p(.)\right]$, $\boldsymbol{\Sigma}= \diag \left(\sigma_1^2(.,.), \ldots, \sigma_p^2(.,.) \right)$, 
such that $\{\mu_i(.), \sigma_i^2(.,.) \}$ is the predictive mean and covariance, respectively, of a GP model of the objective $y_i$.
Note that the predictive distribution of an observation is: 
\begin{equation}\label{eq:F}
 \F(\x) \sim \mathcal{N} \left( \boldsymbol{\mu}(\x), \boldsymbol{\Sigma}\left(\x,\x\right) + \diag(\tau_1^2, \ldots, \tau_p^2) \right).
\end{equation}
%

GPs are commonly limited to a few thousands of design points, due to the cubic cost needed to invert the covariance matrix. 
This can be overcome in several ways, by using inducing points \cite{wilson2015kernel}, or local models \cite{gramacy2015local,rulliere2016nested}; 
see also \cite{Heaton2017} for a comparison or \cite{vzilinskas2016stochastic} for a broader discussion.
In addition, here the full-factorial structure of the design space can potentially be exploited. 
For instance, if evaluated points also have a full-factorial structure, then the covariance matrix can be written as a Kronecker product, 
reducing drastically the computational cost, see e.g., \cite{Plumlee2014a}.

\subsubsection{Sequential design}\label{sec:bo}
Bayesian optimization methods are usually outlined as follows: 
a first set of observations $\{\X_{n_0}, \f_{n_0}\}$ is generated using 
a space-filling design to obtain a first predictive distribution of $\Y(.)$.
Then, observations are performed sequentially by maximizing a so-called
\textit{acquisition function} (or infill criterion) $J(\x)$, that represents the potential
usefulness of a given input $\x$. That is, at step $n \geq n_0$,
\begin{equation}\label{eq:maxJ}
  \x_{n+1} = \arg \max_{\x \in \Xset} J(\x).
\end{equation}

Typically, an acquisition function offers an efficient trade-off between exploration of unsampled regions (high posterior variance) 
and exploitation of promising ones (low posterior mean), and has an analytical expression which makes it inexpensive to evaluate, 
conveniently allowing to use of-the-shelf optimization algorithms to solve Eq.~(\ref{eq:maxJ}).
In unconstrained, noise-free optimization, the canonical choice for $J(\x)$ is the so-called \textit{Expected Improvement} 
\citep[EI,][]{jones1998efficient}, while in the bandit literature (noisy observations),  
the Upper Confidence Bound \citep[UCB,][]{srinivas2012information} can be considered as standard.
Extensions abound in the literature to tackle various optimization problems: see e.g. \citet{wagner2010expected} for multi-objective optimization
or \citet{hernandez2016general} for constrained problems.

\begin{figure}[!hb]
\centering
\includegraphics[trim=5mm 5mm 5mm 10mm, clip, width=.4\textwidth]{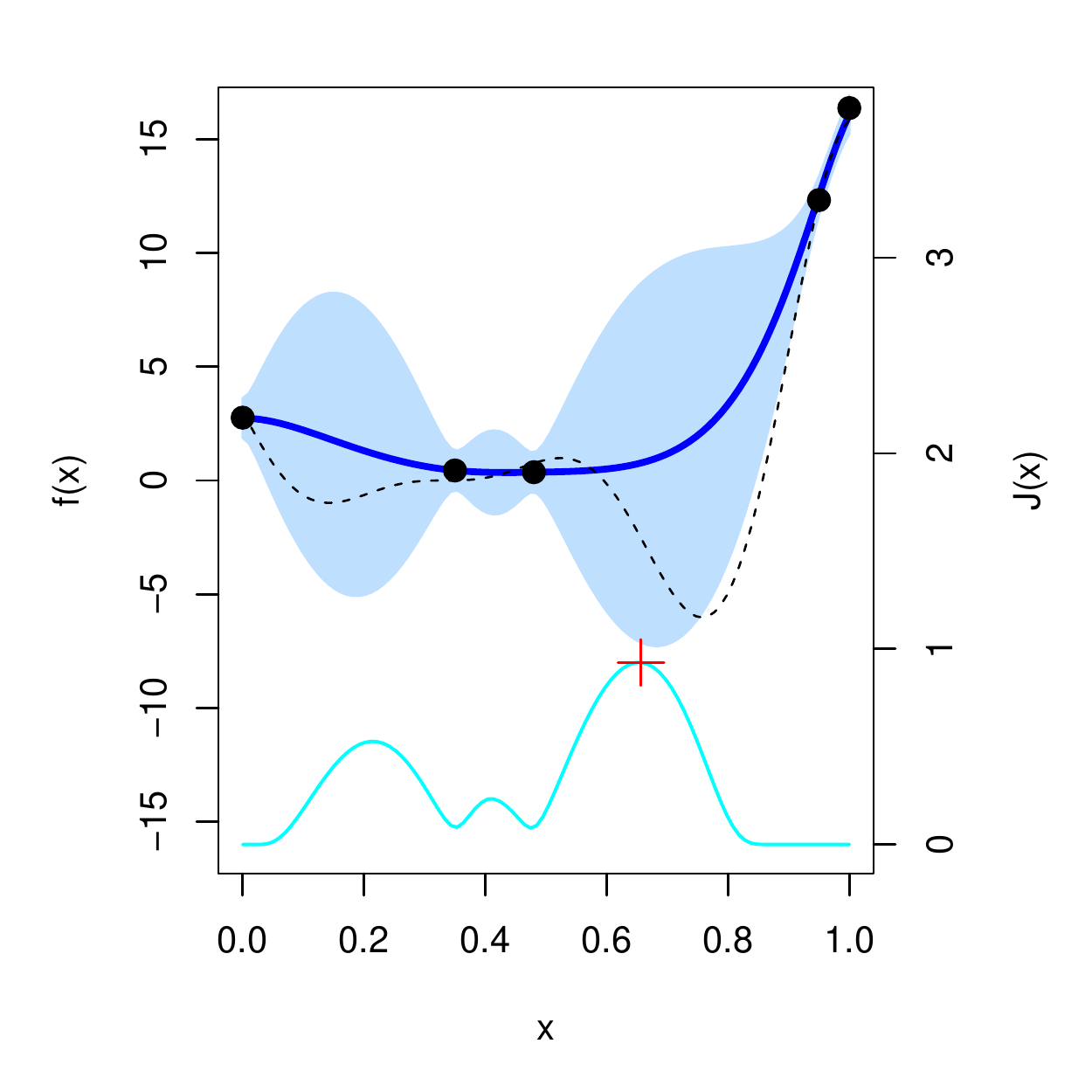}
\includegraphics[trim=5mm 5mm 5mm 10mm, clip, width=.4\textwidth]{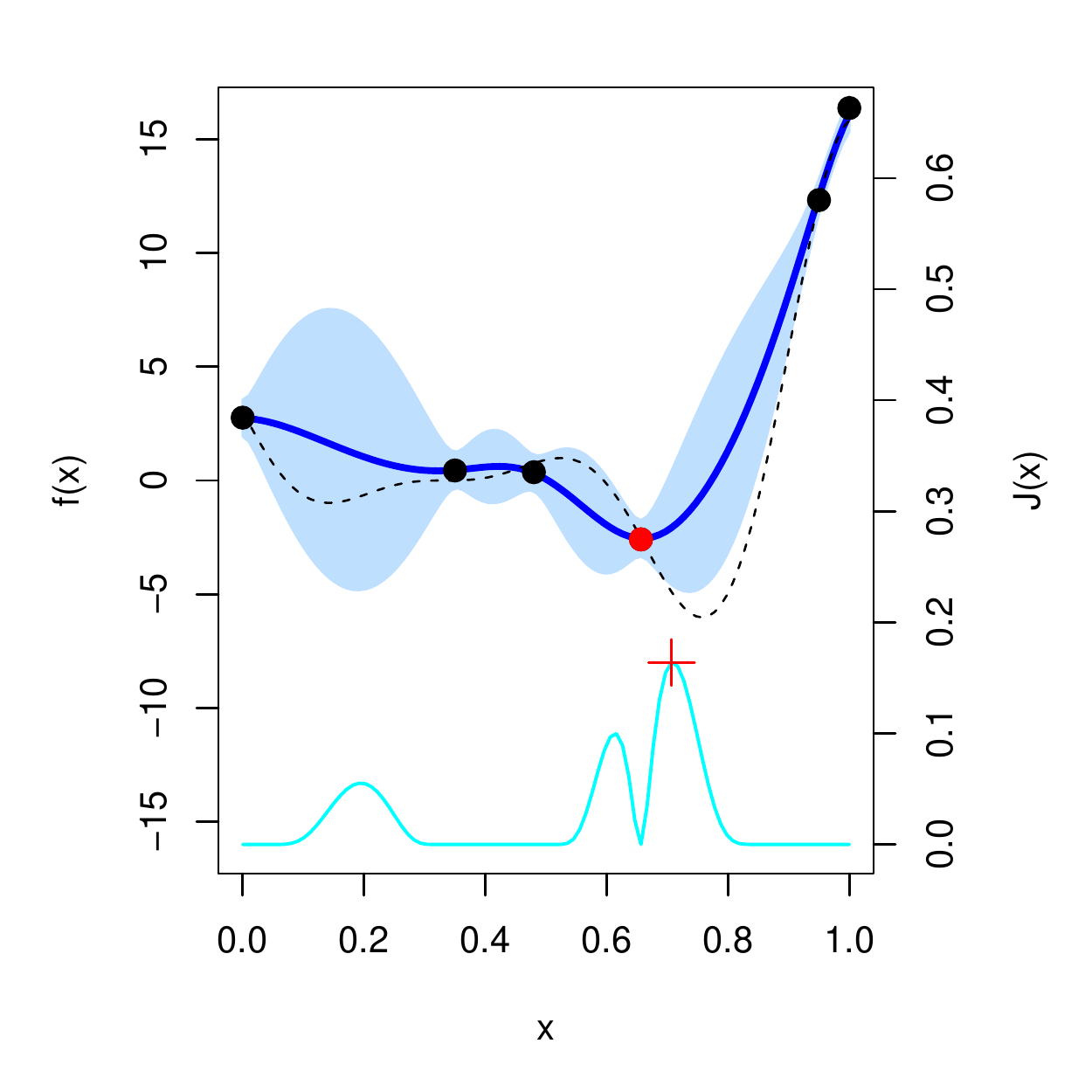}
\includegraphics[width=.18\textwidth]{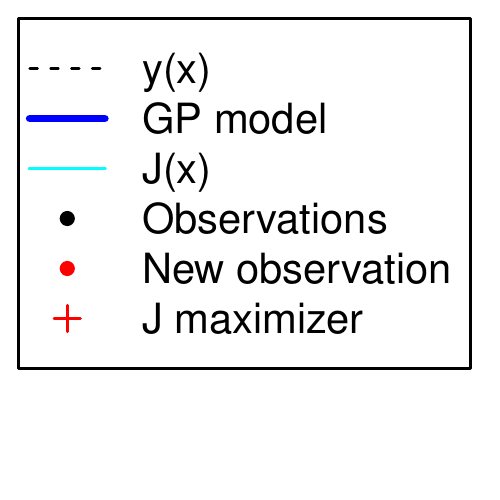}

  \caption{One iteration of Bayesian optimization on a one-dimensional toy problem. A first GP is conditioned on a set of five observations (left), 
  out of which an acquisition function is maximized to find the next observation. Once this observation is performed (right), 
  the GP model and acquisition function are updated, and start pointing towards the optimum $x=0.75$. Note that the acquisition is also large in unexplored regions (around $x=0.2$).} 
\label{fig:EGO1D} 
  \end{figure}

Figure \ref{fig:EGO1D} provides an illustration of the BO principles. 

\section{Acquisition functions for NEP}\label{sec:sequential}
We propose in the following two acquisition functions tailored to solve NEPs, respectively based on the probability of achieving equilibrium 
and on stepwise uncertainty reduction. Both aim at providing an efficient trade-off between exploration and exploitation.

\subsection{Probability of equilibrium}
Given a predictive distribution of $\Y(.)$, 
a first natural metric to consider is the probability of achieving the NE. 
From (\ref{eq:NEdef}), using the notation $\s=(\s_i, \s_{-i})$, this probability writes:
\begin{equation}
\Prob_E(\s) = \Prob \left( \bigcap \limits_{i = 1}^p \left\{ Y_i(\s_i, \s_{-i}) = \min \limits_{\s_i^k \in \sr_i} Y_i(\s_i^k, \s_{-i}) \right\} \right),
\end{equation}
where $\{\s_i^1, \ldots, \s_i^{m_i}\}$ denotes the $m_i$ alternatives in $\sr_i$, and $\s_{-i}$ is fixed to its value in $\s$.

Since our GP model assumes the independence of the posterior distributions of the objectives, we have: 
\begin{equation}
  \Prob_E(\s) =  \prod \limits_{i = 1}^p \Prob \left\{ Y_i(\s) = \min \limits_{\s_i^k \in \sr_i} Y_i(\s_i^k, \s_{-i}) \right\} :=\prod \limits_{i = 1}^p P_i(\s).
\end{equation}

Let us now introduce the notation $\s_i=\s_i^l$ ($1 \leq l \leq m_i$).
As exploited recently by \citet{Chevalier2012} in a multi-point optimization context, 
each $P_i$ can be expressed as
\begin{equation}
 P_i(\s) = \Prob \left( \bigcap\limits_{k \in m_i, k \neq l} \left\{ Y_i(\s_i^l, \s_{-i}) - Y_i(\s_i^k, \s_{-i}) \leq 0 \right\} \right).
\end{equation}
$P_i(\s)$ amounts to compute the cumulative distribution function (CDF) 
of a Gaussian vector of size $q := m_i - 1$:
\begin{equation}
 P_i(\s) = \Prob \left( \mathbf{Z}_i \leq \mathbf{0} \right) = \boldsymbol{\Phi}_{\boldsymbol{\mu}_{Z_i}, \boldsymbol{\Sigma}_{Z_i}}(\mathbf{0}),
\end{equation}
with
\begin{eqnarray*}
 \mathbf{Z}_i &=& \left[Y_i(\s_i^1, \s_{-i}) - Y_i(\s_i^l, \s_{-i}), \ldots, Y_i(\s_i^{l-1}, \s_{-i}) - Y_i(\s_i^l, \s_{-i}), \right.\\
&& \left. Y_i(\s_i^{l+1}, \s_{-i}) - Y_i(\s_i^l, \s_{-i}), \ldots, Y_i(\s_i^{m_i}, \s_{-i}) - Y_i(\s_i^l, \s_{-i}) \right].
\end{eqnarray*}
The mean $\boldsymbol{\mu}_{Z_i}$ and covariance $\boldsymbol{\Sigma}_{Z_i}$ of $\mathbf{Z}_i$ can be expressed as:
\begin{eqnarray*}
 (\mu_{Z_i})_j &=& \mu_i(\x^l_i, \x_{-i}) - \mu_i(\x^j_i, \x_{-i}), \\
 (\Sigma_{Z_i})_{jk} &=& c_i^{ll} + c_i^{jk} - c_i^{lj} - c_i^{lk} \quad \text{ if } k,l \neq j  \text{ and } c_i^{ll}  \text{ otherwise,}
\end{eqnarray*}
with $c_i^{jk} = \sigma_i^2\left( (\x^j_i, \x_{-i}), (\x^k_i, \x_{-i})\right)$. 


Several fast implementations of the multivariate Gaussian CDF are available, for instance in the \texttt{R} packages \texttt{mnormt} 
(for $q < 20$) \citep{Azzalini2016} or, up to $q = 1000$, by Quasi-Monte-Carlo with \texttt{mvtnorm} \citep{Genz2009,Genz2016}. 

Alternatively, this quantity can be computed using Monte-Carlo methods by drawing $R$ samples 
$\yr_i^{(1)}, \ldots, \yr_i^{(R)}$ of $\left[ Y_i(\s_i^1, \s_{-i}), \ldots, Y_i(\s_i^{m_i}, \s_{-i}) \right]$,
to compute
\begin{equation*}
 \hat{P_i}(\s) = \frac{1}{R}\sum_{r=1}^R \mathbbmss{1} \left({\yr_i^{(r)}(\s) = \min \limits_{\s_i^k \in \sr_i} \yr_i^{(r)}(\s_i^k, \s_{-i})}\right),
\end{equation*}
$\mathbbmss{1}(.)$ denoting the indicator function.
This latter approach may be preferred when the number of alternatives $m_i$ is high (say $>20$), 
which makes the CDF evaluation overly expensive while a coarse estimation may be sufficient.
Note that in both cases, a substantial computational speed-up can be achieved by removing
from the $\sr_i$'s the non-critical strategies. This point is discussed in Section \ref{sec:numerical}.

Using $J(\s)=\Prob_E(\s)$ as an acquisition function defines our first sequential sampling strategy.
The strategy is rather intuitive: i.e., sampling at designs most likely to achieve NE.


Still, maximizing $\Prob_E$ is a \textit{myopic} approach (i.e., favoring an immediate reward instead of a long-term one), which are often sub-optimal \citep[see e.g.][and references therein]{Ginsbourger2010,Gonzalez2016}. 
Instead, other authors have advocated the use of an information gain from a new observation instead, see e.g., \citet{villemonteix2009informational,Hennig2012}, which motivated the definition of an alternative
acquisition function, which we describe next.   

\subsection{Stepwise uncertainty reduction}
\emph{Stepwise Uncertainty Reduction} (SUR, also referred to as \textit{information-based approach}) has recently emerged 
as an efficient approach to perform sequential sampling, with successful applications
in optimization \citep{villemonteix2009informational,picheny2014stepwise,hernandez2014predictive,hernandez2016general} 
or uncertainty quantification \citep{bect2012sequential,jala2016sequential}.  
Its principle is to perform a sequence of observations
in order to reduce as quickly as possible an uncertainty measure related to the quantity 
of interest (in the present case: the equilibrium). 

\subsubsection{Acquisition function definition}
Let us first denote by $\Psi(\y)$ the application that associates a NE with a multivariate function. 
In the case of finite games, we have: $\Psi: \Rset^{N \times p} \rightarrow \Rset^p$, for which a pseudo-code is detailed in Algorithm~\ref{alg:nash-comp}.
If we consider the random process $\Y$ (Eq. \ref{eq:Y}) in lieu of the deterministic objective $\y$,
the equilibrium $\Psi(\Y)$ is a random vector of $\Rset^p$ with unknown distribution.
Let $\Gamma$ be a measure of variability (or residual uncertainty) of $\Psi(\Y)$; 
we use here the determinant of its second moment:
\begin{equation}\label{eq:Gamma}
 \Gamma(\Y) = \det \left[\cov \left(\Psi(\Y) \right) \right].
\end{equation}

The SUR strategy aims at reducing $\Gamma$ by adding sequentially observations $\y(\x)$ on
which $\Y$ is conditioned. An ``ideal'' choice of $\x$ would be:
\begin{equation}
 \x_{n+1} = \arg \min_{\x \in \Xset} \Gamma \left[ \Y | \f=\y(\x) \right],
\end{equation}
where $\Y | \f=\y(\x)$ is the process conditioned on the observation $\f$. Since we do not want to evaluate $\y$ for all $\x$ candidates,
we consider the following criterion instead:
\begin{equation}
 J(\x) = \esp_{\F} \left( \Gamma \left[ \Y | \F=\Y(\x) + \boldsymbol{\varepsilon} \right] \right),
\end{equation}
with $\F$ following the posterior distribution (conditioned on the $n$ current observations, Eq.~\ref{eq:F}) and $\esp_{\F}$ denoting the expectation over $\F$.

In practice, computing $J(\x)$ is a complex task, as no closed-form
expression is available. The next subsection is dedicated to this question.

\paragraph{Remark} For simplicity of exposition, we assume here that the equilibrium $\Psi(\Y)$ exists,
which is not guaranteed even if $\Psi(\y)$ does. To avoid this problem, one may consider an extended $\bar \Psi$
function equal to $+\infty \times \mathbb{I}_p$ if there is no equilibrium and to $\Psi$ otherwise,
and in Eq.~\ref{eq:Gamma} use the restriction of $\bar \Psi$ to finite values. 

\subsubsection{Approximation using conditional simulations}
Let us first focus on the measure $\Gamma$ when no new observation is involved.
Due to the strong non-linearity of $\Psi$, no analytical simplification is available, so we
rely on conditional simulations of $\Y$ to evaluate $\Gamma$.

Let $\Yr_1, \ldots, \Yr_M$ be independent draws of $\Y(\Xset)$ (each $\Yr_i \in \Rset^{N \times p}$).
For each draw, the corresponding NE $\Psi(\Yr_i)$ can be computed by exhaustive search. 
We reported to Appendix \ref{app:algNEP} the particular algorithm we used for this step.
The following empirical estimator of $\Gamma( \Y)$ is then available:
\begin{equation*}
 \hat \Gamma \left( \Yr_1, \ldots, \Yr_M \right) = \det \left[ \mathbf{Q}_{\yr}\right],
\end{equation*}
with $\mathbf{Q}_{\yr}$ the sample covariance of $\Psi(\Yr_1), \ldots, \Psi(\Yr_M)$.

Now, let us assume that we evaluate the criterion for a given candidate observation point $\x$.
Let $\fr^1, \ldots, \fr^K$ be independent draws of $\F(\x) = \Y(\x)  + \boldsymbol{\varepsilon} $. For each $\fr^i$, we can condition 
$\Y$ on the event $(\F(\x)=\fr^i)$ in order to generate $\Yr_1|\fr^i, \ldots, \Yr_M|\fr^i$ draws of $\Y | \fr^i$,
from which we can compute the empirical estimator $\hat \Gamma \left( \Yr_1|\fr^i, \ldots, \Yr_M|\fr^i \right)$.
Then, an estimator of $J(\x)$ is obtained using the empirical mean:
\begin{equation*}
 \hat J(\x) = \frac{1}{K}\sum_{i=1}^K \hat \Gamma \left( \Yr_1|\fr^i, \ldots, \Yr_M|\fr^i \right).
\end{equation*}

\subsubsection{Numerical aspects}\label{sec:numerical}
The proposed SUR strategy has a substantial numerical cost, as the criterion requires a double loop for its computation: 
one over the $K$ values of $\fr^i$ and another over the $M$ sample paths.
The two computational bottlenecks are the sample path generations and the searches of equilibria, 
and both are performed in total $K \times M$ times for a single estimation of $J$.
Thankfully, several computational shortcuts allow us to evaluate the criterion efficiently. 

First, we employed the FOXY algorithm (\textit{fast update of conditional simulation ensemble}) as proposed in \citet{chevalier2015fast},
in order to obtain draws of $\Y | \fr^i$ based on a set of draws $\Yr_1, \ldots, \Yr_M$.
In short, a unique set of draws is generated prior to the search of $\x_{n+1}$, which is updated quickly when necessary depending on the pair $(\x, \fr^i)$. 
The expression used are given in Appendix~\ref{app:foxy}, and we refer to \cite{chevalier2015fast} for the detailed algebra and complexity analysis.

Second, we discard points in $\Xset$ that are unlikely to provide information regarding the equilibrium
prior to the search of $\x_{n+1}$, as we detail below. By doing so, we reduce substantially the sample paths size and the dimension 
of each finite game, which drastically reduces the cost. We call $\Xset_\text{sim}$ the retained subset.

Finally, $\hat J$ is evaluated only on a small, promising subset of $\Xset_\text{sim}$. 
We call $\Xset_\text{cand}$ this set.

To select the subsets, we rely on a fast-to-evaluate 
score function $C$, which can be seen as a proxy to the more expensive 
acquisition function. The subset of $\Xset$ is then chosen by 
sampling randomly with probabilities proportional to the scores $C(\Xset)$,
while ensuring that the subset retains a factorial form.
We propose three scores, of increasing complexity and cost, 
which can be interleaved:
\begin{itemize}
\item $C_{\text{target}}$: the simplest score is the posterior density at a target $T_E$ in the objective space, 
for instance the NE of the posterior mean (hence, it requires one NE search). $C_{\text{target}}$ reflects a proximity to an estimate of the NE.
We use this scheme for the first iteration to select $\Xset_\text{sim} \subset \Xset$.
\item $C_{\text{box}}$: once conditional simulations have been performed, the above scheme can be replaced by the probability for a given strategy to fall into the box defined by the extremal values of the 
simulated NE (i.e., $\Psi(\Yr_1), \ldots, \Psi(\Yr_M)$). We use this scheme to select $\Xset_\text{sim} \subset \Xset$ for all the other iterations.
\item $C_{\Prob}$: since $\Prob_E$ is faster (in particular in its Monte Carlo setting with small $R$) than $\hat J(\x)$, it can be used to select $\Xset_\text{cand} \subset \Xset_\text{sim}$.
\end{itemize}
The detailed expressions of $C_{\text{target}}$ and $C_{\text{box}}$ are given in Appendix~\ref{app:C}.
Note that in our experiments, $C_{\text{target}}$ and $C_{\text{box}}$ are also used with the $\Prob_E$ acquisition function.

Last but not least, this framework enjoys savings from parallelization in several ways.
In particular, the searches of NE for each sample $\Yr$ can be readily distributed.

An overview of the full SUR approach is given in pseudo-code in Algorithm \ref{alg:SUR}. Note that 
by construction, SUR does not necessarily sample at the NE, even when it is well-identified. 
Hence, as a post-processing step, the returned NE estimator is the design that maximizes the probability of achieving equilibrium.

\begin{algorithm}[!ht]
\caption{Pseudo-code for the SUR approach}
\begin{algorithmic}[1]
\Require $n_0$, $n_{\max}$, $N_\text{sim}$, $N_\text{cand}$
\State Construct initial design of experiments $\X_{n_0}$
\State Evaluate $\y_{n_0} = \F(\X_{n_0})$
\While{$n \leq n_{\max}$}
\State Train the $p$ GP models on the current design of experiments $\{\X_n, \y_n\}$
\If{$n=n_0$}
\State estimate $T_E = \Psi(\boldsymbol{\mu}(\Xset)$, the NE on the posterior mean; select $\Xset_\text{sim} \subset \Xset$ using $C_{\text{target}}$
\Else 
\State Select $\Xset_\text{sim} \subset \Xset$ using $C_{\text{box}}$
\EndIf
\State Generate $M$ draws $(\Yr_1, \ldots, \Yr_M)$  on $\Xset_\text{sim}$
\State Compute $\Psi(\Yr_1), \ldots, \Psi(\Yr_M)$ (for $C_{\text{box}}$)
\State Select $\Xset_\text{cand} \subset \Xset_\text{sim}$ using $C_{\Prob}$
\State Find $\x_{n+1} = \arg \min_{\x \in \Xset\cand} \hat J(\x)$
\State Evaluate $\y_{n+1} = \F(\x_{n+1})$ and add $\{\x_{n+1}, \y_{n+1}\}$ to the current design of experiments
\EndWhile
\Ensure $\x^* = \arg \min_{\x \in \Xset\cand} \Prob_E(\x)$
\end{algorithmic}
\label{alg:SUR}
\end{algorithm}

\subsubsection{Stopping criterion}
\label{ssec:conver}
We consider in Algorithm~\ref{alg:SUR} a fixed budget, assuming that simulation cost is limited.
However, other natural stopping criteria are available. For $\Prob_E$, one may stop if there is a
strategy for which the probability is high, i.e., $\max \Prob_E(\s) \geq 1-\epsilon$ (with $\epsilon$ close to zero).
For SUR, $\hat J$ is an indicator of the remaining uncertainty, so the algorithm may stop if this uncertainty 
is below a threshold, i.e., $\min J(\s) \leq \epsilon$.

While there always exists a Nash equilibrium for mixed strategy, in the setup
we entertain there may actually be no pure strategy, or several. For $\Prob_E$, this is
completely transparent, with a probability zero for all strategies (no NE) or several
strategies having probability one (multiple NEs). For SUR, with more than one
equilibrium, no change is needed, even though SUR may benefit from using a
clustering method of the simulated NEs and defining local variability instead
of the global $\Gamma$, Eq.~(\ref{eq:Gamma}). The absence of NE on the GP draws ($\Psi(\Yr_i)$)
can be used to detect the absence of NE for the problem at hand.

%
%
%
%
%


\section{NUMERICAL EXPERIMENTS}\label{sec:experiments}
These experiments have been performed in \texttt{R} \citep{R2016} using the \texttt{GPGame} package \citep{picheny2017}, 
which relies on the \texttt{DiceKriging} package \citep{Roustant2012} for the Gaussian process regression part.

We used a fixed-point method as a competitive alternative to compute Nash equilibria, 
in order to assess the efficiency of our approach. 
It is a popular method among the audience who is familiar with gradient-descent optimization algorithms.
The algorithm pseudo-code is given in Algorithm \ref{alg:fixed-point}, and has been implemented in 
\texttt{Scilab} \citep{Scilab2012}.
%

\begin{algorithm}[!ht]
\caption{Pseudo-code for the fixed-point approach \cite{Rubinstein}}
\begin{algorithmic}[1]
\Require $P$: number of players,  $0<\alpha<1$ : relaxation factor, $k_{\max}$ : max iterations
\State Construct initial strategy $\x^{(0)}$
\While{$k \leq k_{\max}$}
\State  Compute in parallel : $\forall i,\ 1\le i \le P, \quad \z^{(k+1)}_{i} = \arg \min_{\x_i \in \Xset_i} J_i(x^{(k)}_{-i} ,x_{i})$
\State Update : $\x^{(k+1)} = \alpha \z^{(k+1)} + (1-\alpha) x^{(k)} $ 
\State \If{$\| \x^{(k+1)} - \x^{(k)}\|$ small enough}  exit \EndIf
\EndWhile
\Ensure For all $i=1...P$, $\x^{*}_{i} = \arg \min_{\x_i \in \Xset_i} J_i(x^{*}_{-i} ,x_{i})$
\end{algorithmic}
\label{alg:fixed-point}
\end{algorithm}

In the following, performance is assessed in terms of numbers of calls to the objective function, hence assuming that the cost of running the black-box model 
largely exceeds the cost of choosing the points. For moderately expensive problems, the choice of algorithm may depend on the budget, as, intuitively, 
the time used to search for a new point may not exceed the time to simulate it. We report in Appendix \ref{app:CPU} the computational times required for our approach
on the three following test problems.

\subsection{A classical multi-objective problem}\label{sec:P1}
We first consider a classical optimization toy problem (P1) as given in \cite{Parr2012}, with two variables and two cost functions, defined as:
\begin{eqnarray*}
 y_1&=&(x_2-5.1(x_1/(2\pi))^2+\frac{5}{\pi}x_1-6)^2 + 10((1-\frac{1}{8\pi})\cos(x_1)+1)\\
 y_2&=&-\sqrt{(10.5-x_1)(x_1+5.5)(x_2+0.5)} - \frac{(x_2 -5.1(x_1/(2\pi))^2-6)^2}{30}\\ &-& \frac{(1-1/(8\pi))\cos(x_1)+1}{3}
\end{eqnarray*}
with $x_1 \in [-5,10]$ and $x_2 \in [0,15]$. Both functions are non-convex.
We set $\x_1 = x_1$ and $\x_2 = x_2$ 
.
The actual NE is attained at $x_1=-3.786$ and $x_2=15$.


Our strategies are parameterized as follow.
$\Xset$ is first discretized over a $31 \times 31$ regular grid.
With such size, there is no need to resort to subsets, so 
we use $\Xset_\text{cand} = \Xset_\text{sim} = \Xset$.
We set the number of draws to $K = M = 20$, which was empirically found as a good trade-off between accuracy and speed.
We use $n_0=6$ initial observations from a Latin hypercube design \citep[LHD, ][]{mckay1979comparison}, and observations are added 
sequentially using both acquisition functions. As a comparison, we ran a standard fixed-point algorithm 
\citep{MR942837} based on finite differences.
This experiment is replicated five times with different initial sets of observations for the GP-based approaches 
and different initial points for the fixed-point algorithm. The results are reported in Table \ref{tab:P1results}.

\begin{figure}[!ht]
\centering
\includegraphics[width=.32\textwidth]{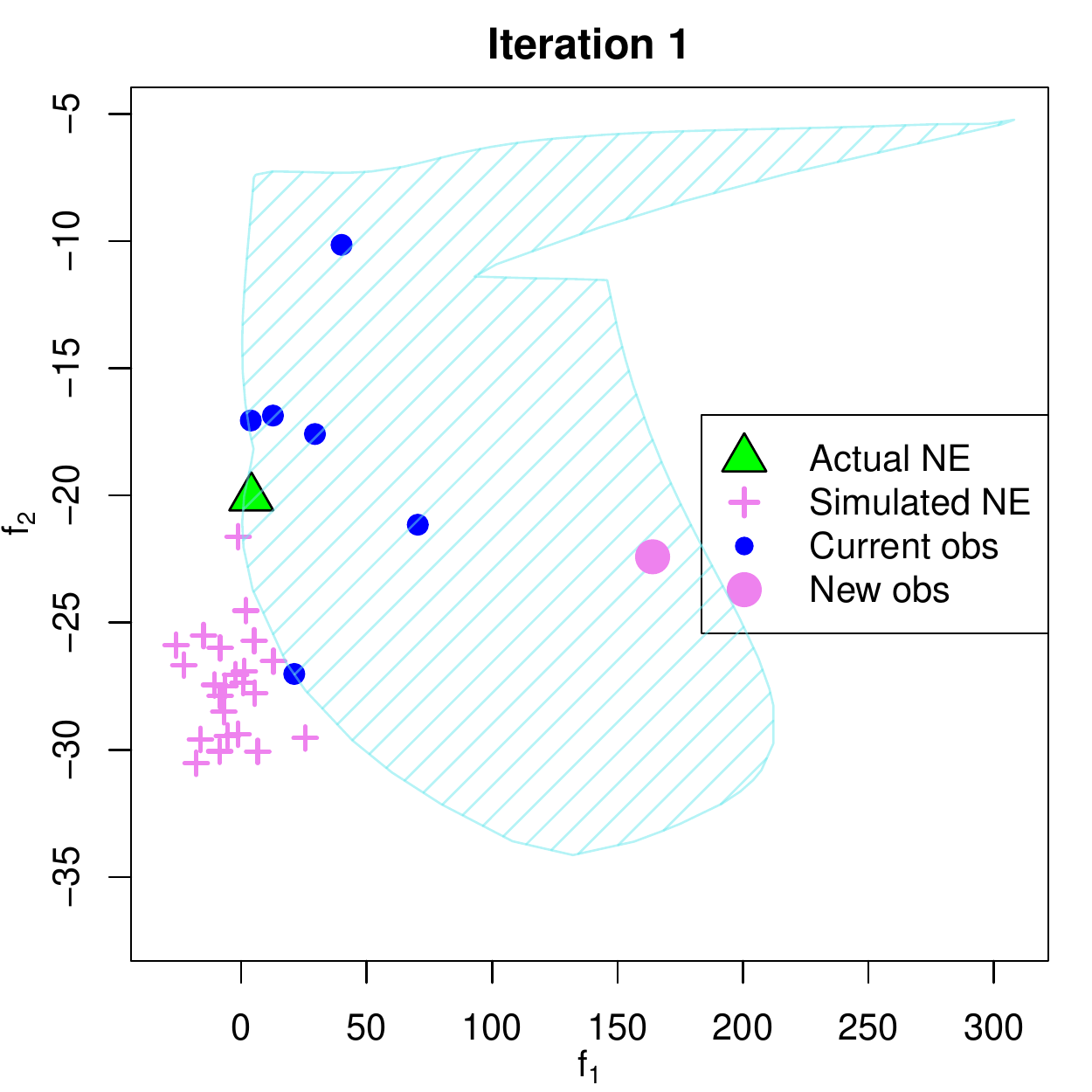}
\includegraphics[width=.32\textwidth]{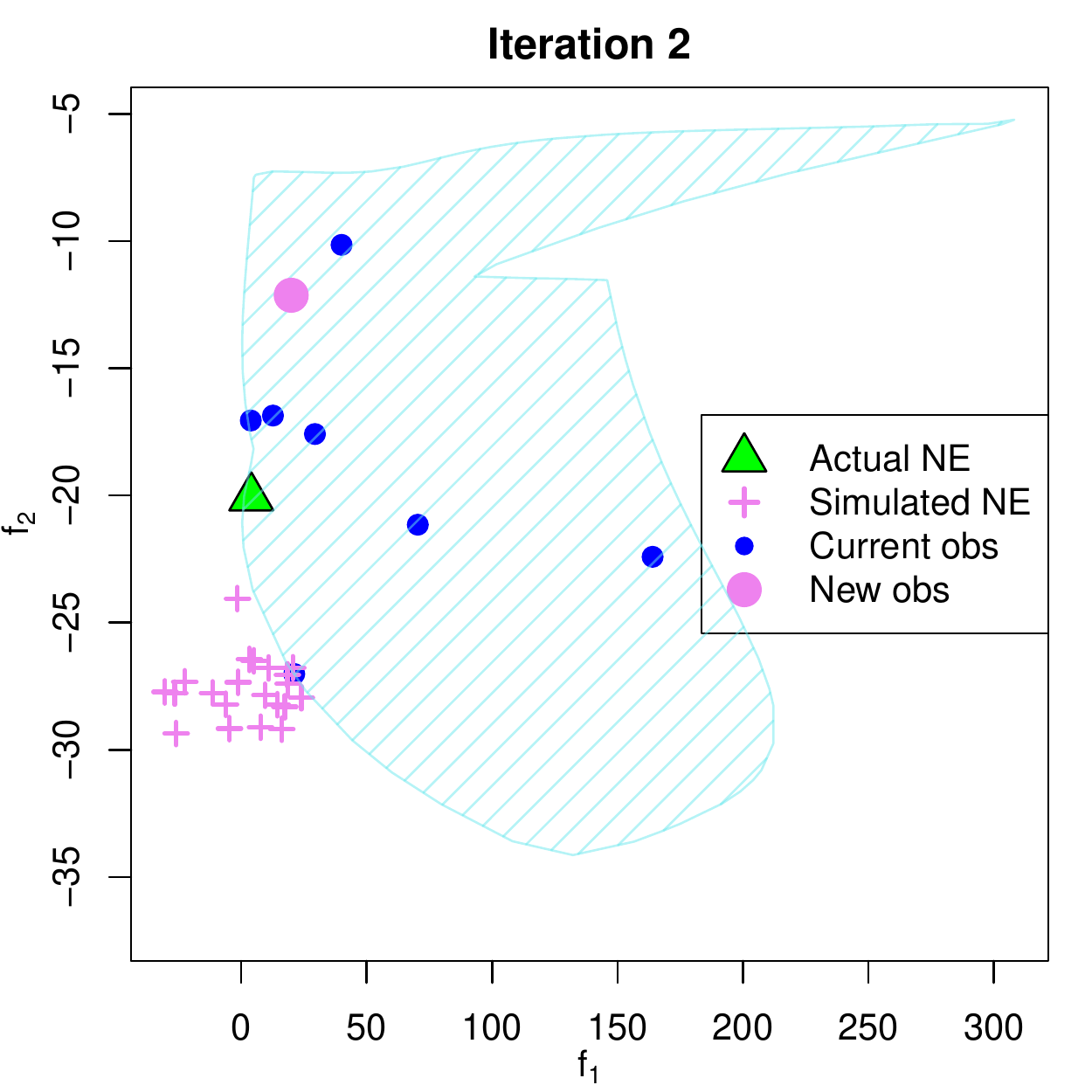}
\includegraphics[width=.32\textwidth]{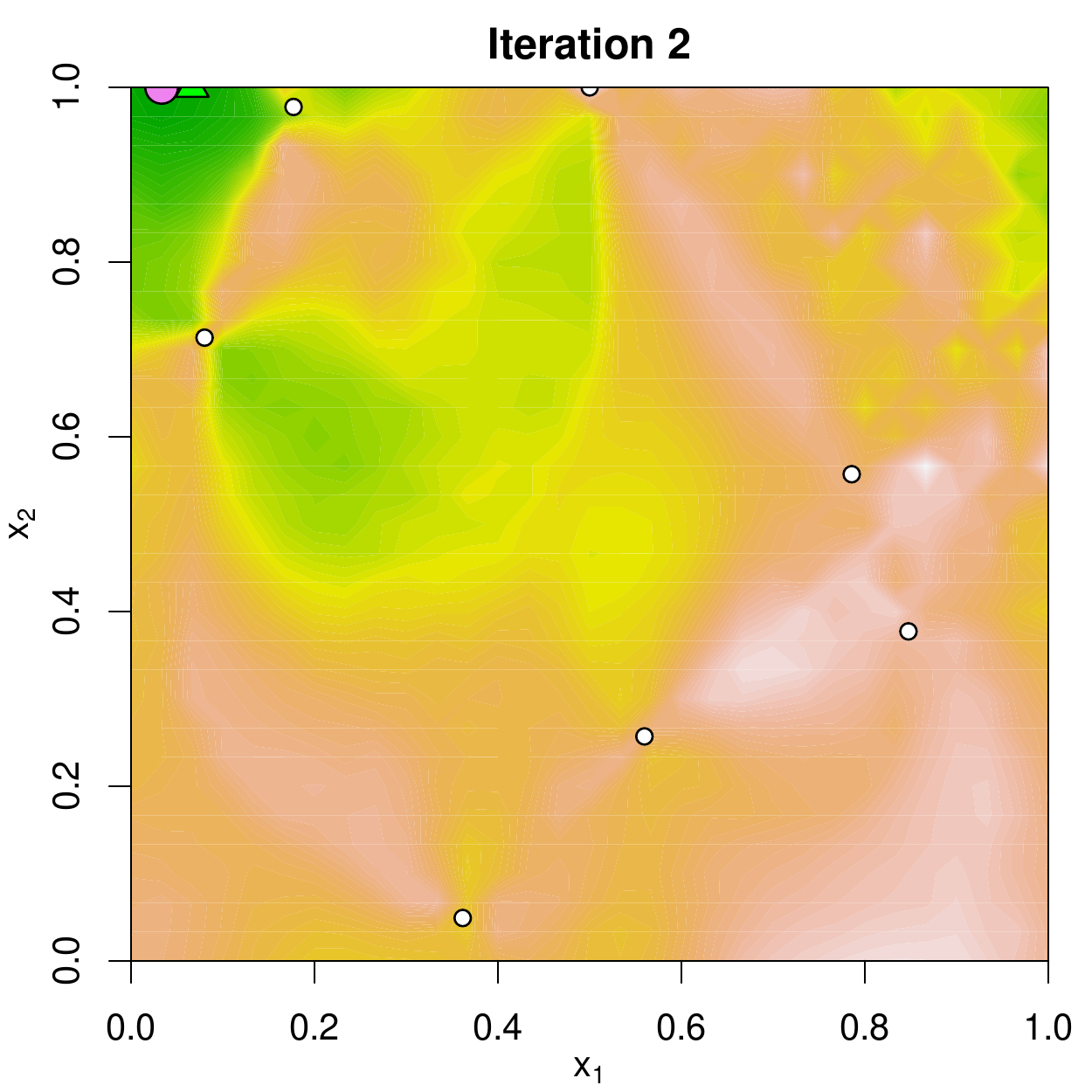}
\includegraphics[width=.32\textwidth]{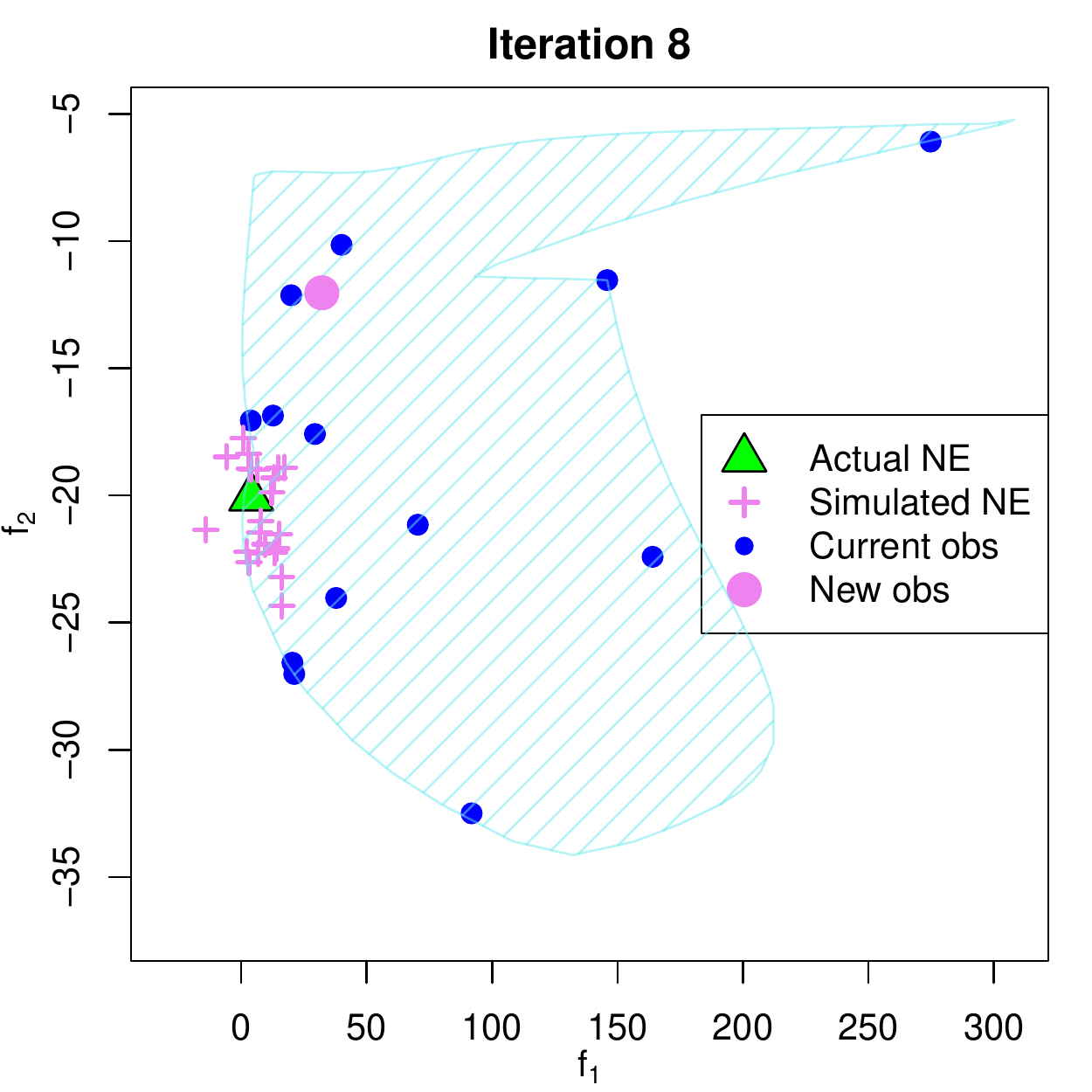}
\includegraphics[width=.32\textwidth]{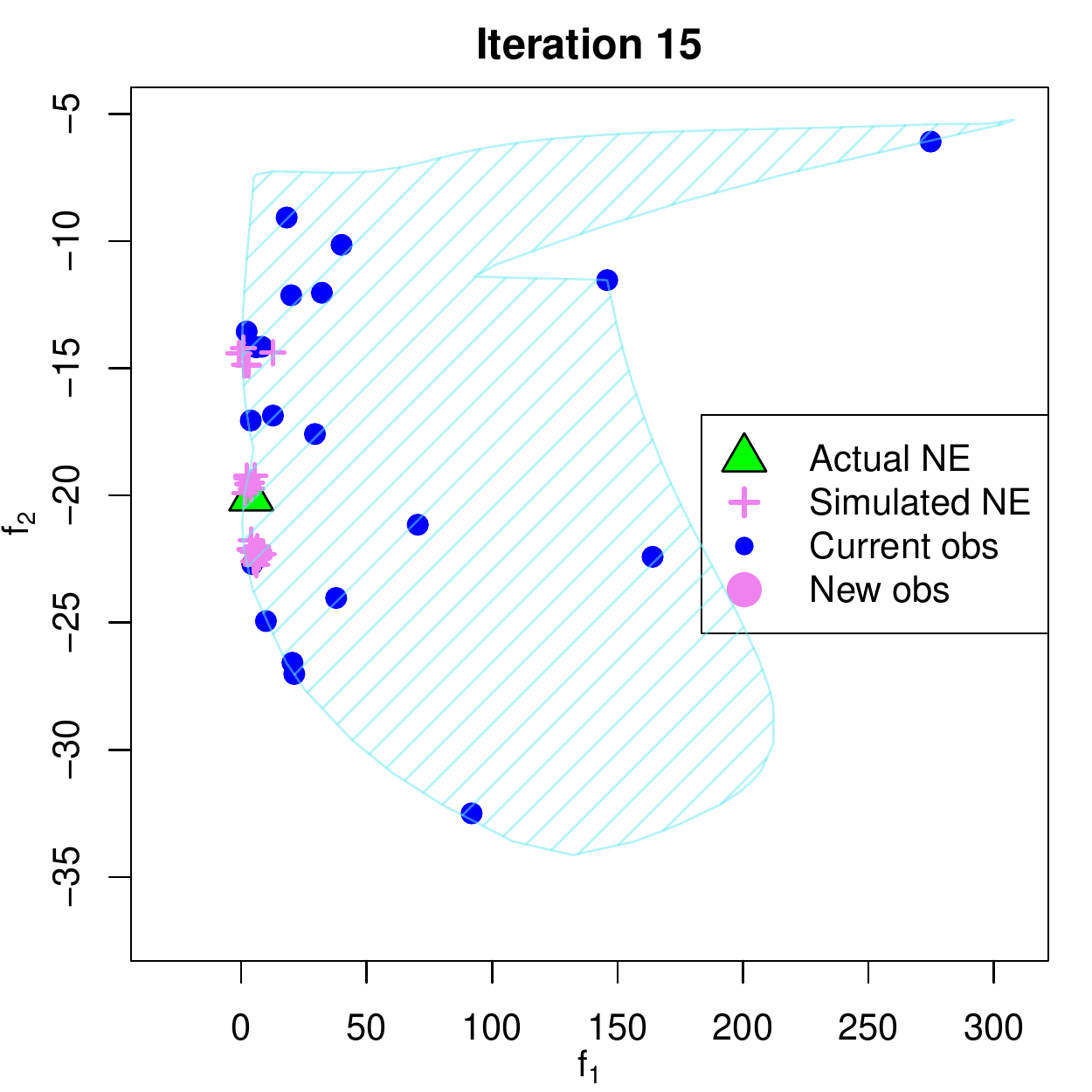}
\includegraphics[width=.32\textwidth]{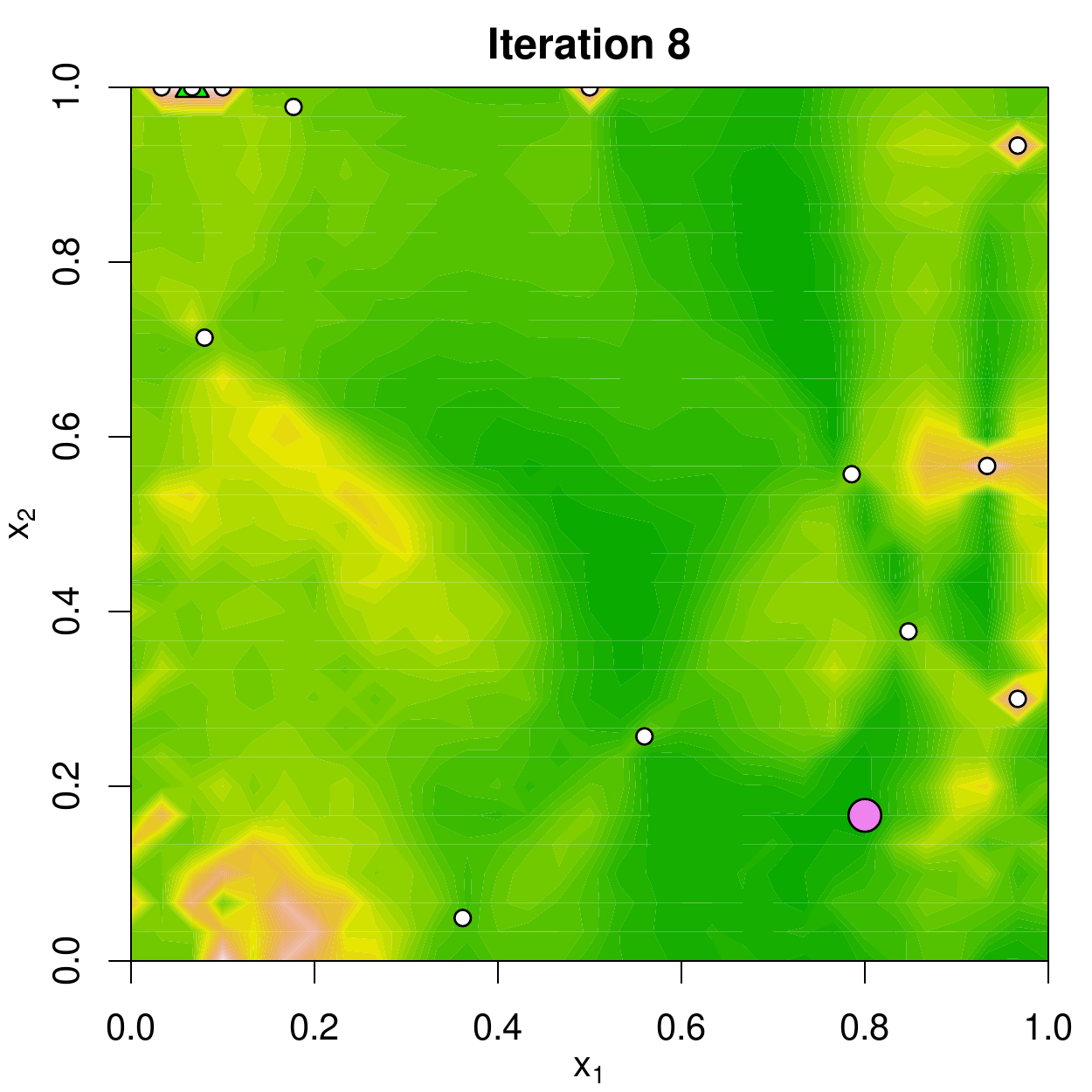}
  \caption{Four iterations of SUR for (P1). Left and middle: observations and simulated NEs in the objective space. The hatched area represents the image of $\Xset$ by $\y$. 
  Right: level sets of the SUR criterion value in the variable space (green is better).}
  \label{fig:P1}
\end{figure}

In addition, Figure \ref{fig:P1} provides some illustration for a single run. In the initial state (top left),
the simulated NEs form a cloud in the region of the actual NE. A first point is added, although far from the NE, 
that impacts the form of the cloud (top, middle). The infill criterion surface (Figure \ref{fig:P1}, top right) 
then targets the top left corner, which offers the best compromise between exploration and exploitation. 
After adding 7 points (bottom left) the cloud is smaller and centered on the actual NE.
As the NE is quite well-identified, the infill criterion surface (Figure \ref{fig:P1}, bottom right) 
is now relatively flat and mostly indicates regions that have not been explored yet.
After 14 additions (bottom, middle), all simulated NE but two concentrate on the actual NE. 
The observed values have been added around the actual NE,
but not only, which indicates that some exploration steps (such as iteration 8) have been performed. 


\begin{table}[!ht]
 \begin{center}
\begin{tabular}{ccc}
Strategy & Evaluations required & Success rate \\
\hline
$\Prob_E$ & 9--10 & 5/5 \\
SUR & 8--14 & 5/5 \\
Fixed point & 200--1000 & 3/5
\end{tabular}
\end{center}
\caption{P1 convergence results.}
\label{tab:P1results}
\end{table}

Both GP approaches consistently find the NE very quickly: the worst run required 14 cost evaluations (that is, 8 infill points).
In contrast, the classical fixed-point algorithm required hundreds of evaluations, which is only marginally better
than an exhaustive search on the 961-points grid. Besides, two out of five runs converged to the stationary point $(1,1)$,
which is not a NE. On this example, $\Prob_E$ performed slightly better than SUR.

\subsection{An open loop differential game}\label{ssec:differential}
Differential games model a huge variety of competitive interactions, in social behavior, economics, biology among many others \citep[predator-prey, pursuit-evasion games and so on, see][]{MR0210469}. As a toy-model, let us
consider $p$ players who compete to drive a dynamic process to their preferred final state. 
Let denote $T>0$ the final time, and $t\in [0,T]$ the time variable. Then we consider the following simple process, which state $\z(t) \in \R^2$ obeys the first order dynamics:
\begin{eqnarray}
\left\{
\begin{array}{lll} \label{edostate}
\dot{\z}(t) &=& \mathbf{v}_0 + \sum_{1\le i \le p} \alpha_i(t) \s_i(t) \\
\z(0) &=& \z_0 
\end{array}
\right.
\end{eqnarray}
The parameter $\alpha_i(t) = e^{-\theta_i t}$ models the lifetime range of each player's influence on the process ($\theta_i$ is a measure of player (i)'s preference for the present). 
The time-dependent action $\s_i(t) \in \Rset^2$ is the player (i)'s strategy, which we restrict to the (finite-dimensional) space of spline functions of a given degree $\kappa-1$, that is:
%
\begin{equation*}
  \s_i(t) = \left[\begin{array}{cc} \sum_{1\le k \le \kappa} a^{(i)}_{k} \times B_k(t/T)\\ \sum_{1\le k \le \kappa} b^{(i)}_{k} \times B_k(t/T) \end{array} \right],
\end{equation*}
where $\left( B_k \right)_{1\le k \le \kappa}$ is the spline basis.

Now, the decision variables for player (i) is the array 
 $(a^{(i)}_{1}, \ldots, a^{(i)}_{\kappa}, b^{(i)}_{1}, \ldots, b^{(i)}_{\kappa})$ (the spline coefficients),
and $\x \in \Rset^{\kappa \times 2 \times p}$. 
We used $\kappa=1$ (constant splines, two decision variables per player) and 
$\kappa=2$ (linear splines, four decision variables per player).

All players have their own preferred final states $\z_T^{i}\in \R^2$, and a limited energy to dispense in playing with their own action.
The cost $y_i$ of player (i) is then the following :
\begin{equation}
y_i(\s_i, \s_{-i}) = \dfrac{1}{2} \| \z(T) - \z_T^{i} \|_{\R^2}^2 +  \dfrac{1}{2} \| \s_i \|_{L^2(0,T)}^2.
\end{equation}
The game considered here is an open loop differential game (the decisions do not depend on the state), and belongs to the larger class of dynamic Nash games.

\begin{figure}[!hb]
\centering
\includegraphics[width=70mm]{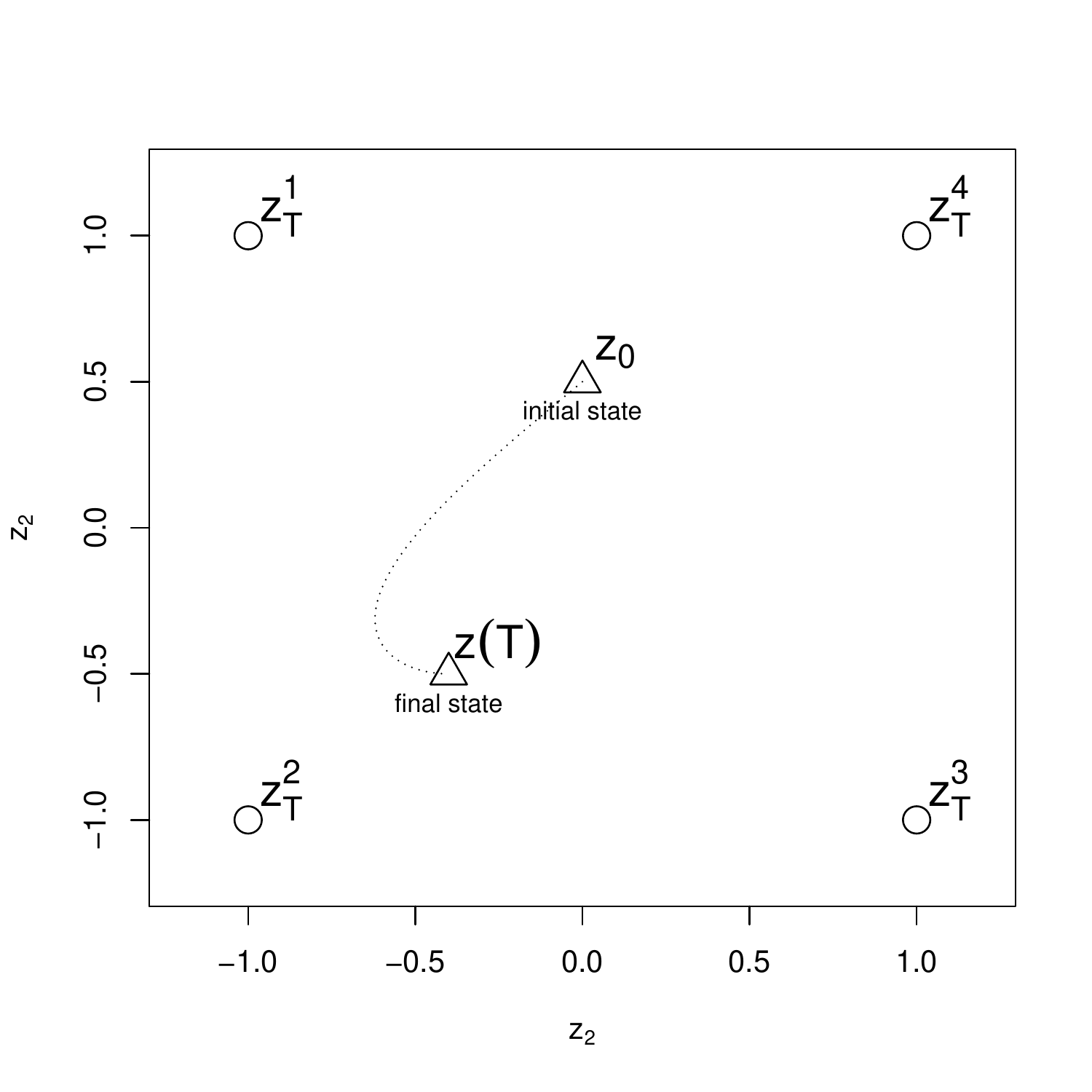}
\caption{The differential game setting: The process starts at the initial state $\z_0$ and travels during a time $T$ to reach the state $\z(T)$. 
Players targets are $\z_T^{(i)}$. The final state at Nash equilibrium is $\z_{NE}$.}
\label{fig:DGsetting}
\end{figure}

When the $\alpha_i$'s do not depend on $i$, and the points $\z_T^{i}$ are located on some -any- circle, 
then the Nash solution of the game puts the final state $\z(T)$ on the center of the circle 
(provided $T$ is large enough to let the state reach this center starting from $z_0$). 
In our setup, we have $p = 4$ players, with targets positioned on the four corners of the $[-1,1]^2$ square, as illustrated by Figure \ref{fig:DGsetting}.
The state equation (\ref{edostate}) is solved by means of an explicit Euler numerical scheme, with $T$ set to 4, and 40 time steps. 
As for problem parameters, we chose $\mathbf{v}_0 = \mathbf{0}$, $\z_0 = (0,0.5)$ (non-central initial position)  and $\boldsymbol{\theta} = (\theta_i) = (0.25, 0, 0.5, 0)$ (heterogeneous time preferences), 
which makes the search of NE non-trivial.
 The results "Fixed point" presented in Table \ref{tab:diffresultsK1} are performed by means of a popular fixed-point algorithm, as described in Algorithm-\ref{alg:fixed-point}.

The initial design space is set as $[-6,6]^d$ (with $d=8$ or $d=16$) and discretized as follow.
For each set of design variables belonging to a player, we first generate a 17-point LHD. 
For $\kappa=1$, this means four LHD in the spaces $(x_1, x_2)$, $(x_3, x_4)$ and so on.
Then, we take all the combinations of strategies, ending with a total of $\text{Card}(\Xset)= N=17^4=83,521$ possible strategies.

In both cases, we followed Algorithm \ref{alg:SUR} with $n_0 = 80$ or $160$ initial design points and a maximum budget of $n_{\max} = 160$ or $320$ evaluations (depending on the dimension).
We chose $\text{Card}(\Xset_\text{sim})= N_\text{sim}=1296$ simulation points and $\text{Card}(\Xset\cand)= N\cand=256$ candidates.
For the number of draws to compute the SUR criterion, we chose $K =M = 20$. 

A fixed-point algorithm based on finite differences is also ran, and
the experiment is replicated five times with different algorithm initializations.
Note that, here, the actual NE is unknown beforehand. However, since all runs (including the fixed-point algorithm ones) converged to the same point,
we assume that it is the actual NE.

We show the results in Table \ref{tab:diffresultsK1}.
For the lower dimension problem ($d=8$), all runs found the solution with less than 100 evaluations (that is, $80$ initial points plus $20$ infills). 
This represents about 1\% of the total number of possible strategies.
In this case, SUR appears as slightly more efficient than $\Prob_E$.
For the higher dimension problem ($d=16$), more evaluations are needed, yet all runs converge 
with less than 240 evaluations. As a comparison, on both cases the fixed-point algorithm is more than 60 (resp. 20) times more expensive. 

\begin{table}[!t]
 \begin{center}
\begin{tabular}{ccc}
Configuration & Strategy & Evaluations required \\
\hline
$d=8 \quad (\kappa=1)$ & $\Prob_E$ & 83--95 \\
$d=8 \quad (\kappa=1)$  & SUR & 81--88 \\
$d=8 \quad (\kappa=1)$  & Fixed point & 3000--5000 \\
$d=16 \quad (\kappa=2) $ & $\Prob_E$ & 196--221 \\
$d=16 \quad (\kappa=2) $ & SUR & 208--232 \\
$d=16 \quad (\kappa=2) $ & Fixed point & 5000--7000
\end{tabular}
\end{center}
\caption{Differential game convergence results.}
\label{tab:diffresultsK1}
\end{table}
\def\ds{\displaystyle}
\def\R{\mathbb{R}}
\def\Li{H^{-\frac{1}{2}}(\Gamma_i)}
\def\Hi{H^{\frac{1}{2}}(\Gamma_i)}
\def\HO{H^{1}(\Omega)}
\def\Lc{H^{-\frac{1}{2}}(\Gamma_c)}
\def\Hc{H^{\frac{1}{2}}(\Gamma_c)}
\def\Ldi{L^{2}(\Gamma_i)}
\def\Ldc{L^{2}(\Gamma_c)}

\subsection{PDE-constrained example: data completion}\label{sec:pde}
\subsubsection{Problem description}
We address here the class of problems known as data completion or data recovery problems. 
Let be $\Omega$ a bounded open domain in $\R^d$ ($d=2,\, 3$) with a sufficiently smooth boundary 
$\partial \Omega$ composed of two connected disjoint components  $\Gamma_c$ and $\Gamma_i$, 
with the latter being inaccessible to boundary measurements. 
For details, see \cite{HabbalSICON2013} whence the present example is excerpt. 

Let us focus, for illustration, on the particular case of steady state heat equation. The problem is formulated in terms of the following elliptic Cauchy problem : 
\begin{eqnarray}
\label{CP}
\left\{
\begin{array}{rcccl}
\nabla.(\lambda\nabla u) & = & 0 & \mathrm{ in } & \Omega \\
 u & = & \varphi & \mathrm{ on } & \Gamma_c \\
 \lambda\nabla u . \nu & = & \Phi & \mathrm{ on } & \Gamma_c \\
\end{array}\right.
\end{eqnarray}
The data to be recovered, or missing data, are $u_{|\Gamma_i}$ and $\lambda\nabla u. \nu_{|\Gamma_i}$, which are determined as soon as one knows $u$ in the whole $\Omega$. 
The parameters $\lambda$, $\varphi$ and $\Phi$ are given functions, $\nu$ is the unit outward normal vector on the
boundary. The Dirichlet data $\varphi$ and the Neumann data $\Phi$ are the so-called
Cauchy data, which are known on the accessible part $\Gamma_c$ of the boundary
$\partial \Omega$ and the unknown field $u$ is the Cauchy solution. 
Completion/Cauchy problems are known to be severely ill-posed (Hadamard's), and computationally challenging.
 
Let us assume that  $(\Phi, \varphi) \in \Lc\times\Hc$ where $\Lc$ resp. $\Hc$ are the Sobolev spaces of Neumann resp. Dirichlet traces of functions in the Sobolev space $\HO$ 
\citep[see e.g.][]{adams2003sobolev}. 
For given  $\eta \in \Li$ and $\zeta \in \Hi$, let us define $u_1(\eta)$ and
$u_2(\zeta)$ as the unique solutions in $\HO$ of the following elliptic boundary value problems :
\begin{eqnarray}
\label{OP}
\hspace{-1.5cm}(SP1)\left\{
\begin{array}{rcccl}
\nabla.(\lambda\nabla u_1) & = & 0 & \mathrm{ in } & \Omega \\
 u_1 & = & \varphi & \mathrm{ on } & \Gamma_c \\
 \lambda\nabla u_1 . \nu & = & \eta & \mathrm{ on } & \Gamma_i \\
\end{array}\right.
& \,\,
(SP2)\left\{
\begin{array}{rcccl}
\nabla.(\lambda\nabla u_2) & = & 0 & \mathrm{ in } & \Omega \\
 u_2 & = & \zeta & \mathrm{ on } & \Gamma_i \\
 \lambda\nabla u_2 . \nu & = & \Phi & \mathrm{ on } & \Gamma_c \\
\end{array}\right.
\end{eqnarray}

Let us define the following two costs : for any $\eta \in \Li$ and $\zeta \in \Hi$, 
\begin{eqnarray}
J_1(\eta, \zeta) &=& \ \frac{1}{2}\| \lambda\nabla u_1 . \nu - \Phi\|_{\Lc}^2  +\ \frac{\alpha}{2}\|u_1 - u_2\|_{\Hi}^2,\label{DefJ1}\\
J_2(\eta, \zeta) &=& \ \frac{1}{2}\| u_2 - \varphi\|_{\Ldc}^2  +\ \frac{\alpha}{2}\|u_1 - u_2\|_{\Hi}^2,\label{DefJ2}
\end{eqnarray}
%
where 
$\alpha$ is a given positive parameter (e.g. $\alpha=1$). 

Let us remark that, for the problem is severely ill posed,  the classical descent algorithms for the minimization of the cost $J_1+J_2$ with respect to the couple of variables $(\eta, \zeta)$ do not converge.
Besides, the solution is extremely sensitive to small variations of $\phi$ and $\Phi$. This originally motivated the above-described game formulation.

The fields $u_1=u_1(\eta)$ and $u_2=u_2(\zeta)$ are aiming at the fulfillment of a possibly antagonistic goals, namely minimizing the Neumann gap \mbox{$\|\lambda\nabla u_1 . \nu - \Phi\|_{\Lc}$} and the Dirichlet gap $\|u_2 - \varphi\|_{\Ldc}$. 
This antagonism is intimately related to Hadamard's ill-posedness character of the Cauchy problem, and rises as soon as one requires that $u_1$ and $u_2$ coincide, which is exactly what the coupling term $\|u_1 - u_2\|_{\Ldi}$ is for. 
Thus, one may think of an iterative process which minimizes in a smart fashion the three terms, namely Neumann-Dirichlet-Coupling terms.

From a game theory perspective, one may define two players, one associated with the strategy variable $\eta$ and cost $y_1=J_1$ and the second with the variable $\zeta$ and cost $y_2=J_2$,
each trying to minimize its cost in a non-cooperative fashion.
The fact that each player controls only his own strategy, 
while there is a strong dependence of each player's cost on the joint strategies $(\eta, \zeta)$ 
justifies the use of the game theory framework (and terminology), 
a natural setting which may be used to formulate the negotiation between these two costs. 
%
%

\begin{theorem}\citep{HabbalSICON2013} There always exists a unique Nash equilibrium $(\eta^*,\zeta^*)\in\Li\times\Hi$, and when the Cauchy problem has a solution $u$, 
then $u_1(\eta^*) = u_2(\zeta^*) = u$, and $(\eta^*,\zeta^*)$ are the missing data, namely  $\eta^*=\lambda\nabla u . \nu_{|\Gamma_i}$ and $\zeta^*=u_{|\Gamma_i}$.
\end{theorem}


\subsubsection{Noisy data and random Nash equilibrium}
In a realistic situation, the fields $\varphi$ and $\Phi$ may be ``polluted'' by noise.
\citet{HabbalSICON2013} showed that the Nash equilibrium is stable with respect to small perturbations, 
and that the perturbed equilibrium converges strongly to the unperturbed one when the noise tends to zero.
However, they did not consider directly the random Nash game:
\begin{eqnarray}
\left\{
\begin{array}{ll}
 \min_{\eta} \esp[J_1(\eta, \zeta, \varphi^\epsilon, \Phi^\epsilon)] \\
 \min_{\zeta} \esp[J_2(\eta, \zeta, \varphi^\epsilon, \Phi^\epsilon)],
\end{array}\right.
\end{eqnarray}
where $\varphi^\epsilon$ and $\Phi^\epsilon$ are perturbed values of $\varphi$ and $\Phi$.

Addressing this problem with a classical fixed-point algorithm \textit{\`a la} \cite{MR942837} would require,
for each trial pair $(\eta, \zeta)$, repeated evaluations of $J_1$ and $J_2$ for many different values of 
$\varphi^\epsilon$ and $\Phi^\epsilon$, which would prove extremely intensive computationally.

\subsubsection{Implementation and experimental setup}

\begin{figure}[!hb]
\centering
\includegraphics[width=70mm]{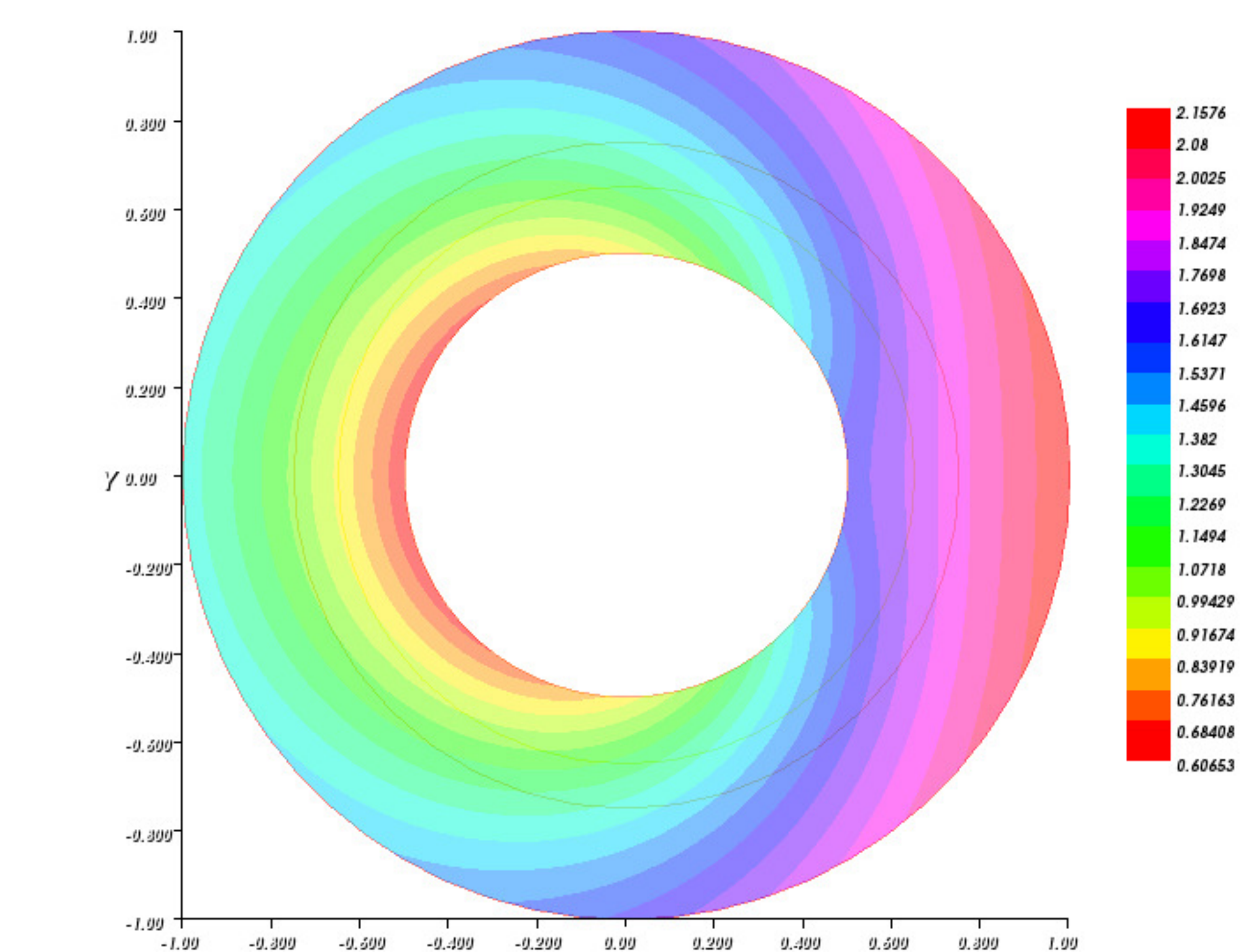} 
\caption{The domain is an annular structure, the outer circle is the accessible boundary while the inner circle is the inaccessible one. The plot shows the isovalues of the solution $u_1$ at convergence.}
\label{fig:completion_sol}
\end{figure}

The physical domain $\Omega$ is taken as a 2D annular structure. 
The accessible boundary $\Gamma_c$ is the outer circle, with a ray $R_c = 1$ and the inaccessible boundary $\Gamma_i$ is 
the inner circle with a ray $R_i = 0.5 $, see Figure \ref{fig:completion_sol}. 
The conductivity coefficient is $\lambda=1$, the flux $\Phi=0$ and the heat field $\varphi$ is built from an exact known solution (an academic $u(x,y)=\exp(x)\cos(y)$ used for validation issues).

We use FreemFem++ \cite{hecht2010freefem++} to develop our finite element (FE) solvers. 
The FE computations are performed with a $P1$-triangular mesh yielding $1,088$ degrees of freedom, 
the outer and inner boundaries being discretized each with $60$ finite element nodes. 
From now on, we use the same notations as above for functions to refer to their finite element approximations (values at FE nodes).

As in Section \ref{ssec:differential}, to reduce the dimensionality of the problem, 
the $\eta$ and $\zeta$ being originally vectors of size 60 (the number of nodes at the inner boundary), 
we interpolate the underlying functions with -natural- splines with $\kappa=8$ coefficients for each quantity, 
resulting with a decision space of size $d=16$ (instead of the original 120).
Each coefficient is bounded between -3 and 3, allowing a large variety of spline shapes.


Since the Neumann condition involving $\Phi$ is known to be the most sensitive to noise, 
we perturb only this term with an additive noise:
\begin{equation*}
 \Phi^\epsilon = \Phi_0 + \epsilon(\xi),
\end{equation*}
with $\epsilon(\xi)$ a white noise uniformly distributed between $-0.25$ and $0.25$, 
thus resulting in a randomized vector of dimension 60 (the number of FE nodes on the outer boundary).
The resulting variability on $J_1(\eta, \zeta, \Phi^\epsilon)$ and $J_2(\eta, \zeta, \Phi^\epsilon)$
strongly depends on the value of $(\eta, \zeta)$ and can be very large (with coefficients of variation, $y_i/\tau_i$, from 5\% to over 100\%).

The original continuous space of the spline coefficients is discretized by generating first two $1000 \times 8$ LHD
and taking all the combinations between the two designs, ending with space of size $N = 10^6$ potential strategies.
The Bayesian optimization strategy is run with $n_0 =50$ initial points (chosen by space-filling design) and 150 infill points. 
Since the noise is very heteroskedastic, we use five repetitions for each trial pair $(\eta, \zeta)$ in order 
to obtain a rough estimate of the noise variance $\tau_i$, and we take the mean over those five repetitions as our observations ($f_i$'s). 
This results with a total budget of $(50+150) \times 5=1,000$ model runs.	
Let us notice that computing a \emph{deterministic} NE with fixed-point methods requires about five times this budget on this problem.
We followed Algorithm \ref{alg:SUR} with the SUR criterion, $K =M = 20$, selecting $\text{Card}(\Xset_\text{sim})= N_\text{sim}=1,296$ simulation points and $\text{Card}(\Xset\cand)= N\cand=256$ candidates.

\subsubsection{Results}
\begin{figure}[!hb]
  \centering
  \includegraphics[width=\textwidth]{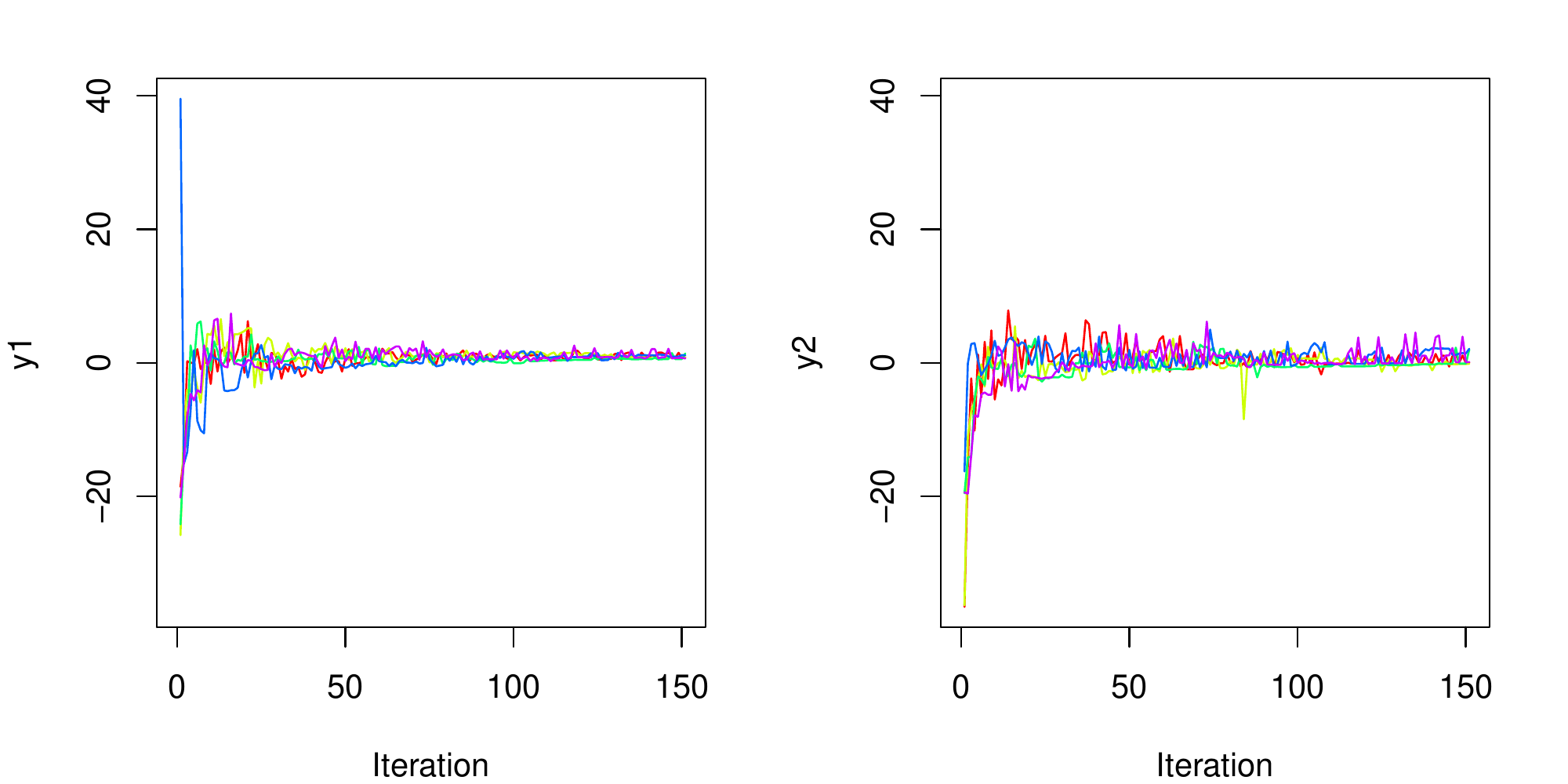} 
  \caption{Convergence of the estimated Nash equilibrium for five algorithm runs. Each iteration corresponds to five calls of the FreeFem++ model.}
  \label{fig:res_BHP}
\end{figure}

The algorithm is run five times with different initial points to assess its robustness. 
The results on the test problem are presented in Figure \ref{fig:res_BHP}. 
In the absence of reference (\textit{ground truth} to be compared to), 
we show the evolution of the two objectives functions during optimization.
All five runs identify very quickly (after 20 iterations, which corresponds to 350 FE evaluations) a similar estimation of the equilibrium, close 
to the equilibrium of the noiseless problem (which is actually known to be $(0,0)$, \citet{HabbalSICON2013}). 
However, it fails at converging finely to a single, stable value, in particular for $y_2$. 
This is not surprising and can largely be attributed to the difficulty of the problem (with respect to the noise level and dimensionality). 
Hybrid approaches (using fixed-point strategies for a final local convergence) or sophisticated sampling strategies 
(increasing the number of repeated simulations gradually with the iterations) may be used to solve this issue 
(although at the price of a considerably higher computational budget).

\section{CONCLUDING COMMENTS}\label{sec:conclusion}
We have proposed here a novel approach to solve stochastic or deterministic Nash games with drastically limited budgets of evaluations based on GP regression, taking the form of a Bayesian optimization algorithm.
Experiments on challenging synthetic problems demonstrate the potential of this approach compared to classical, derivative-based algorithms.
%

On the test problems, the two acquisition functions performed similarly well. $\Prob_E$ has the benefit of not relying on conditional simulation paths, which makes it simpler
to implement and less computationally intensive in most cases. Still, the SUR approach has several decisive advantages; in particular,
it does not actually require the new observations to belong to the grid $\Xset_\text{sim}$, such that it could be optimized continuously. 
Moreover, it lays the groundwork for many extensions that may be pursued in future work.


First, SUR strategies are well-suited to allow 
selecting batches of points instead of only one, a key feature in distributed computer experiments \citep{Chevalier2012}.
Second, other games and equilibria may be considered: the versatility of the SUR approach
may allow its transposition to other frameworks. 
In particular, mixed equilibria, which existence are always guaranteed on discrete problems, could be addressed by
using $\Psi$ functions that return discrete probability measures.
Generalized NEPs \citep{facchinei2010generalized} could be tackled by building on existing works on Bayesian optimization with constraints 
\citep[see e.g.,][and references therein]{hernandez2016general}.

Finally, this work could also be extended to continuous domains, for which related convergence properties may be investigated 
in light of recent advances on SUR approaches \citep{bect2016supermartingale}.

\appendix

\section{Handling conditional simulations}\label{app:foxy}
We detail here how we generate the draws of $\Y | \Fr_i$ to compute $\hat J(\x)$ in practice. 
We employ the FOXY (\textit{fast update of conditional simulation ensemble}) algorithm proposed in \citet{chevalier2015fast},
as detailed below.

Let $\Yr_1, \ldots, \Yr_M$ be independent draws of $\Y\left(\Xset \right)$ (each $\Yr_i \in \Rset^{N \times p}$), 
generated using the posterior Gaussian distribution of Eq.~(\ref{eq:Y}), and 
$\Fr_1, \ldots, \Fr_K$ independent (of each other and of the $\Yr_i$'s) draws of $\Y(\x) + \boldsymbol{\varepsilon}$ 
from the posterior Gaussian distribution of Eq.~(\ref{eq:F}).
As shown in \citet{chevalier2015fast}, draws of $\Y | \Fr_i$ can be obtained efficiently from $\Yr_1, \ldots, \Yr_M$ using:
\begin{eqnarray}
 \yr_j^{(i)} | \fr_k^{(i)} &=& \yr_j^{(i)} +  \boldsymbol{\lambda}^{(i)}(\x) \left( \fr_k^{(i)} - \yr_j^{(i)}(\x) \right),
\end{eqnarray}
with $1 \leq i \leq p$, $1 \leq j \leq M$, $1 \leq k \leq K$ and
\begin{eqnarray*}
\boldsymbol{\lambda}^{(i)}(\x) = \frac{\mathbf{k}_n^{(i)}(\x, \Xset)}{\mathbf{k}_n^{(i)}(\x, \x)}.
\end{eqnarray*}
Notice that $\boldsymbol{\lambda}^{(i)}(\x)$ may only be computed once for all $\yr_j^{(i)}(\x)$.

\section{$C(\x)$ formulae}\label{app:C}
For a given target $T_E \in \Rset^p$ and $\x \in \Xset$:
\begin{equation}
 C_{\text{target}}(\x) = \prod_{i=1}^p \phi\left(\frac{T_{Ei} - \mu_i(\x)}{\sigma_i(\x)} \right),
\end{equation}
with $\phi$ the probability density function of the standard Gaussian variable.

Let $T_L \in \Rset^p$ and $T_U \in \Rset^p$ such that $\forall 1 \leq i \leq p, T_{Li} < T_{Ui}$ define a box in the objective space.
Defining $\boldsymbol{\Psi} = \left[\Psi(\Yr_1), \ldots, \Psi(\Yr_M) \right]$ the $p \times M$ matrix of simulated NE, we use:
\begin{equation*}
 \forall 1 \leq i \leq p \qquad T_{Li} = \min \boldsymbol{\Psi}_{i, 1 \ldots M} \quad \text{ and } \quad T_{Ui} = \max \boldsymbol{\Psi}_{i, 1 \ldots M}.
\end{equation*}
Then, the probability to belong to the box is:
\begin{equation}
 C_{\text{box}}(\x) = \prod_{i=1}^p \left[ \Phi\left(\frac{T_{Ui} - \mu_i(\x)}{\sigma_i(\x)} \right) - \Phi\left(\frac{\mu_i(\x) - T_{Li}}{\sigma_i(\x)} \right) \right].
\end{equation}

\section{Solving NEP on GP draws}\label{app:algNEP}

We detail here a simple algorithm to extract Nash equilibria from GP draws.

\begin{algorithm}[!ht]
\caption{Pseudo-code for Nash equilibria extraction}
\begin{algorithmic}[1]
\Require $P$: number of players, $\mathcal{Y}$: draw of $\Y(\Xset)$ of size $N \times P$, $I = \left\{1, \dots, N \right\}$
\For{$1 \leq i \leq N$}
  \If{$i \in I$}
  \For{$1 \leq j \leq P$}
    \State Find $K = \left\{1 \leq k  \leq N, \x^k_{-j} = \x^i_{-j} \right\}$
    \State Find $l^* = \min \limits_{1 \leq l \leq |K|} \mathcal{Y}(\x^l)_j$
    \For{$1 \leq l \leq |K|$}
      \If{$\mathcal{Y}(\x^l)_j > \mathcal{Y}(\x^{l^*})_j$}
        \State $I \leftarrow I \setminus l$ 
      \EndIf
    \EndFor
  \EndFor
  \EndIf
\EndFor
\Ensure $I$
\end{algorithmic}
\label{alg:nash-comp}
\end{algorithm}

\section{Computational time}\label{app:CPU}
We report here the computational time required to perform a single iteration of our algorithm for each of the three examples 
(not including the time required to run the simulation itself).
Experiments were run on an Intel\textregistered Core\textsuperscript{TM} i7-5600U CPU at 2.60GHz with 4 $\times$ 8GB of RAM.

\begin{table}
 \begin{tabular}{ccc}
 Case / Criterion         &    Pnash   &     SUR \\
          \hline\\
Case \ref{sec:P1}    &     4s      &       20s\\
Case \ref{ssec:differential} ($\kappa=1$)  &  13s     &      40s\\
Case \ref{ssec:differential} ($\kappa=2$)  &  28s      &     82s\\
Case \ref{sec:pde}   &   9s     &        12s
 \end{tabular}
 \caption{Average CPU times required for one iteration of the GP-based algorithm on the different test problems.}
\end{table}



\begin{thebibliography}{64}
\expandafter\ifx\csname natexlab\endcsname\relax\def\natexlab#1{#1}\fi

\bibitem[{Adams \& Fournier(2003)}]{adams2003sobolev}
\textsc{Adams, R.~A.} \& \textsc{Fournier, J.~J.} (2003).
\newblock \textit{Sobolev spaces}, vol. 140.
\newblock Academic press.

\bibitem[{{\'A}lvarez et~al.(2011){\'A}lvarez, Rosasco \&
  Lawrence}]{Alvarez2011}
\textsc{{\'A}lvarez, M.~A.}, \textsc{Rosasco, L.} \& \textsc{Lawrence, N.~D.}
  (2011).
\newblock Kernels for vector-valued functions: A review.
\newblock \textit{Foundations and Trends in Machine Learning} \textbf{4},
  195--266.

\bibitem[{Azzalini \& Genz(2016)}]{Azzalini2016}
\textsc{Azzalini, A.} \& \textsc{Genz, A.} (2016).
\newblock \textit{The {R} package \texttt{mnormt}: The multivariate normal and
  $t$ distributions (version 1.5-4)}.

\bibitem[{Ba{\c{s}}ar(1987)}]{MR942837}
\textsc{Ba{\c{s}}ar, T.} (1987).
\newblock Relaxation techniques and asynchronous algorithms for on-line
  computation of noncooperative equilibria.
\newblock \textit{J. Econom. Dynam. Control} \textbf{11}, 531--549.

\bibitem[{Bect et~al.(2016)Bect, Bachoc \&
  Ginsbourger}]{bect2016supermartingale}
\textsc{Bect, J.}, \textsc{Bachoc, F.} \& \textsc{Ginsbourger, D.} (2016).
\newblock A supermartingale approach to gaussian process based sequential
  design of experiments.
\newblock \textit{arXiv preprint arXiv:1608.01118} .

\bibitem[{Bect et~al.(2012)Bect, Ginsbourger, Li, Picheny \&
  Vazquez}]{bect2012sequential}
\textsc{Bect, J.}, \textsc{Ginsbourger, D.}, \textsc{Li, L.}, \textsc{Picheny,
  V.} \& \textsc{Vazquez, E.} (2012).
\newblock Sequential design of computer experiments for the estimation of a
  probability of failure.
\newblock \textit{Statistics and Computing} \textbf{22}, 773--793.

\bibitem[{Brown et~al.(2015)Brown, Ganzfried \& Sandholm}]{Brown2015}
\textsc{Brown, N.}, \textsc{Ganzfried, S.} \& \textsc{Sandholm, T.} (2015).
\newblock Hierarchical abstraction, distributed equilibrium computation, and
  post-processing, with application to a champion no-limit {T}exas hold'em
  agent.
\newblock In \textit{Proceedings of the 2015 International Conference on
  Autonomous Agents and Multiagent Systems}.

\bibitem[{Chevalier et~al.(2015)Chevalier, Emery \&
  Ginsbourger}]{chevalier2015fast}
\textsc{Chevalier, C.}, \textsc{Emery, X.} \& \textsc{Ginsbourger, D.} (2015).
\newblock Fast update of conditional simulation ensembles.
\newblock \textit{Mathematical Geosciences} \textbf{47}, 771--789.

\bibitem[{Chevalier \& Ginsbourger(2013)}]{Chevalier2012}
\textsc{Chevalier, C.} \& \textsc{Ginsbourger, D.} (2013).
\newblock Fast computation of the multi-points expected improvement with
  applications in batch selection.
\newblock In \textit{Learning and Intelligent Optimization}. Springer, pp.
  59--69.

\bibitem[{Cressie(1993)}]{Cressie1993}
\textsc{Cressie, N.} (1993).
\newblock Statistics for spatial data: Wiley series in probability and
  statistics .

\bibitem[{Dorsch et~al.(2013)Dorsch, Jongen \& Shikhman}]{dorsch2013structure}
\textsc{Dorsch, D.}, \textsc{Jongen, H.~T.} \& \textsc{Shikhman, V.} (2013).
\newblock On structure and computation of generalized nash equilibria.
\newblock \textit{SIAM Journal on Optimization} \textbf{23}, 452--474.

\bibitem[{Facchinei \& Kanzow(2010)}]{facchinei2010generalized}
\textsc{Facchinei, F.} \& \textsc{Kanzow, C.} (2010).
\newblock Generalized nash equilibrium problems.
\newblock \textit{Annals of Operations Research} \textbf{175}, 177--211.

\bibitem[{Fleuret \& Geman(1999)}]{fleuret1999graded}
\textsc{Fleuret, F.} \& \textsc{Geman, D.} (1999).
\newblock Graded learning for object detection.
\newblock In \textit{Proceedings of the workshop on Statistical and
  Computational Theories of Vision of the IEEE international conference on
  Computer Vision and Pattern Recognition (CVPR/SCTV)}, vol.~2.

\bibitem[{Friedman(1972)}]{friedman1972stochastic}
\textsc{Friedman, A.} (1972).
\newblock Stochastic differential games.
\newblock \textit{Journal of differential equations} \textbf{11}, 79--108.

\bibitem[{Games(2016)}]{games2016lenient}
\textsc{Games, I.-L. S.~C.} (2016).
\newblock Lenient learning in independent-learner stochastic cooperative games.
\newblock \textit{Journal of Machine Learning Research} \textbf{17}, 1--42.

\bibitem[{Garivier et~al.(2016)Garivier, Kaufmann \&
  Koolen}]{garivier2016maximin}
\textsc{Garivier, A.}, \textsc{Kaufmann, E.} \& \textsc{Koolen, W.~M.} (2016).
\newblock Maximin action identification: A new bandit framework for games.
\newblock In \textit{29th Annual Conference on Learning Theory}.

\bibitem[{Genz \& Bretz(2009)}]{Genz2009}
\textsc{Genz, A.} \& \textsc{Bretz, F.} (2009).
\newblock \textit{Computation of Multivariate Normal and t Probabilities}.
\newblock Lecture Notes in Statistics. Heidelberg: Springer-Verlag.

\bibitem[{Genz et~al.(2016)Genz, Bretz, Miwa, Mi, Leisch, Scheipl \&
  Hothorn}]{Genz2016}
\textsc{Genz, A.}, \textsc{Bretz, F.}, \textsc{Miwa, T.}, \textsc{Mi, X.},
  \textsc{Leisch, F.}, \textsc{Scheipl, F.} \& \textsc{Hothorn, T.} (2016).
\newblock \textit{{mvtnorm}: Multivariate Normal and t Distributions}.
\newblock R package version 1.0-5.

\bibitem[{Gibbons(1992)}]{Gibbons-Game-92}
\textsc{Gibbons, R.} (1992).
\newblock \textit{Game Theory for Applied Economists}.
\newblock Princeton, NJ: Princeton University Press.

\bibitem[{Ginsbourger \& Le~Riche(2010)}]{Ginsbourger2010}
\textsc{Ginsbourger, D.} \& \textsc{Le~Riche, R.} (2010).
\newblock Towards {G}aussian process-based optimization with finite time
  horizon.
\newblock In \textit{mODa 9--Advances in Model-Oriented Design and Analysis}.
  Springer, pp. 89--96.

\bibitem[{Gonzalez et~al.(2016)Gonzalez, Osborne \& Lawrence}]{Gonzalez2016}
\textsc{Gonzalez, J.}, \textsc{Osborne, M.} \& \textsc{Lawrence, N.} (2016).
\newblock Glasses: Relieving the myopia of {B}ayesian optimisation.
\newblock In \textit{Proceedings of the 19th International Conference on
  Artificial Intelligence and Statistics}.

\bibitem[{Gramacy \& Apley(2015)}]{gramacy2015local}
\textsc{Gramacy, R.~B.} \& \textsc{Apley, D.~W.} (2015).
\newblock Local gaussian process approximation for large computer experiments.
\newblock \textit{Journal of Computational and Graphical Statistics}
  \textbf{24}, 561--578.

\bibitem[{Gramacy \& Ludkovski(2015)}]{Gramacy2015}
\textsc{Gramacy, R.~B.} \& \textsc{Ludkovski, M.} (2015).
\newblock Sequential design for optimal stopping problems.
\newblock \textit{SIAM Journal on Financial Mathematics} \textbf{6}, 748--775.

\bibitem[{Habbal \& Kallel(2013)}]{HabbalSICON2013}
\textsc{Habbal, A.} \& \textsc{Kallel, M.} (2013).
\newblock Neumann-{D}irichlet {N}ash strategies for the solution of elliptic
  {C}auchy problems.
\newblock \textit{SIAM J. Control Optim.} \textbf{51}, 4066--4083.

\bibitem[{Habbal et~al.(2004)Habbal, Petersson \& Thellner}]{Habbal2004-A1}
\textsc{Habbal, A.}, \textsc{Petersson, J.} \& \textsc{Thellner, M.} (2004).
\newblock Multidisciplinary topology optimization solved as a {N}ash game.
\newblock \textit{Int. J. Numer. Meth. Engng} \textbf{61}, 949--963.

\bibitem[{Harsanyi(1973)}]{harsanyi1973games}
\textsc{Harsanyi, J.~C.} (1973).
\newblock Games with randomly disturbed payoffs: A new rationale for
  mixed-strategy equilibrium points.
\newblock \textit{International Journal of Game Theory} \textbf{2}, 1--23.

\bibitem[{Heaton et~al.(2017)Heaton, Datta, Finley, Furrer, Guhaniyogi, Gerber,
  Gramacy, Hammerling, Katzfuss, Lindgren et~al.}]{Heaton2017}
\textsc{Heaton, M.~J.}, \textsc{Datta, A.}, \textsc{Finley, A.},
  \textsc{Furrer, R.}, \textsc{Guhaniyogi, R.}, \textsc{Gerber, F.},
  \textsc{Gramacy, R.~B.}, \textsc{Hammerling, D.}, \textsc{Katzfuss, M.},
  \textsc{Lindgren, F.} et~al. (2017).
\newblock Methods for analyzing large spatial data: A review and comparison.
\newblock \textit{arXiv preprint arXiv:1710.05013} .

\bibitem[{Hecht et~al.(2010)Hecht, Pironneau, Le~Hyaric \&
  Ohtsuka}]{hecht2010freefem++}
\textsc{Hecht, F.}, \textsc{Pironneau, O.}, \textsc{Le~Hyaric, A.} \&
  \textsc{Ohtsuka, K.} (2010).
\newblock Freefem++ v. 2.11.
\newblock \textit{User?s Manual. University of Paris} \textbf{6}.

\bibitem[{Hennig \& Schuler(2012)}]{Hennig2012}
\textsc{Hennig, P.} \& \textsc{Schuler, C.~J.} (2012).
\newblock Entropy search for information-efficient global optimization.
\newblock \textit{The Journal of Machine Learning Research} \textbf{13},
  1809--1837.

\bibitem[{Hern{\'a}ndez-Lobato et~al.(2016)Hern{\'a}ndez-Lobato, Gelbart,
  Adams, Hoffman \& Ghahramani}]{hernandez2016general}
\textsc{Hern{\'a}ndez-Lobato, J.~M.}, \textsc{Gelbart, M.~A.}, \textsc{Adams,
  R.~P.}, \textsc{Hoffman, M.~W.} \& \textsc{Ghahramani, Z.} (2016).
\newblock A general framework for constrained bayesian optimization using
  information-based search.
\newblock \textit{Journal of Machine Learning Research} \textbf{17}, 1--53.

\bibitem[{Hern{\'a}ndez-Lobato et~al.(2014)Hern{\'a}ndez-Lobato, Hoffman \&
  Ghahramani}]{hernandez2014predictive}
\textsc{Hern{\'a}ndez-Lobato, J.~M.}, \textsc{Hoffman, M.~W.} \&
  \textsc{Ghahramani, Z.} (2014).
\newblock Predictive entropy search for efficient global optimization of
  black-box functions.
\newblock In \textit{Advances in neural information processing systems}.

\bibitem[{Hu \& Wellman(2003)}]{hu2003nash}
\textsc{Hu, J.} \& \textsc{Wellman, M.~P.} (2003).
\newblock Nash q-learning for general-sum stochastic games.
\newblock \textit{Journal of Machine learning research} \textbf{4}, 1039--1069.

\bibitem[{Isaacs(1965)}]{MR0210469}
\textsc{Isaacs, R.} (1965).
\newblock \textit{Differential games. {A} mathematical theory with applications
  to warfare and pursuit, control and optimization}.
\newblock John Wiley \& Sons, Inc., New York-London-Sydney.

\bibitem[{Jala et~al.(2016)Jala, L{\'e}vy-Leduc, Moulines, Conil \&
  Wiart}]{jala2016sequential}
\textsc{Jala, M.}, \textsc{L{\'e}vy-Leduc, C.}, \textsc{Moulines, {\'E}.},
  \textsc{Conil, E.} \& \textsc{Wiart, J.} (2016).
\newblock Sequential design of computer experiments for the assessment of fetus
  exposure to electromagnetic fields.
\newblock \textit{Technometrics} \textbf{58}, 30--42.

\bibitem[{Johanson \& Bowling(2009)}]{Johanson2009}
\textsc{Johanson, M.} \& \textsc{Bowling, M.~H.} (2009).
\newblock Data biased robust counter strategies.
\newblock In \textit{Proceedings of the Twelfth International Conference on
  Artificial Intelligence and Statistics (AISTATS)}.

\bibitem[{Jones et~al.(1998)Jones, Schonlau \& Welch}]{jones1998efficient}
\textsc{Jones, D.~R.}, \textsc{Schonlau, M.} \& \textsc{Welch, W.~J.} (1998).
\newblock Efficient global optimization of expensive black-box functions.
\newblock \textit{Journal of Global optimization} \textbf{13}, 455--492.

\bibitem[{Kanzow \& Steck(2016)}]{kanzow2016augmented}
\textsc{Kanzow, C.} \& \textsc{Steck, D.} (2016).
\newblock Augmented lagrangian methods for the solution of generalized nash
  equilibrium problems.
\newblock \textit{SIAM Journal on Optimization} \textbf{26}, 2034--2058.

\bibitem[{Lanctot et~al.(2012)Lanctot, Burch, Zinkevich, Bowling \&
  Gibson}]{Lanctot2012}
\textsc{Lanctot, M.}, \textsc{Burch, N.}, \textsc{Zinkevich, M.},
  \textsc{Bowling, M.} \& \textsc{Gibson, R.~G.} (2012).
\newblock No-regret learning in extensive-form games with imperfect recall.
\newblock In \textit{Proceedings of the 29th International Conference on
  Machine Learning (ICML-12)}.

\bibitem[{{L}e\'on et~al.(2014){L}e\'on, {P}ape, D\'esid\'eri, Alfano \&
  Costes}]{JAD2014-HELICO}
\textsc{{L}e\'on, E.~R.}, \textsc{{P}ape, A.~L.}, \textsc{D\'esid\'eri, J.-A.},
  \textsc{Alfano, D.} \& \textsc{Costes, M.} (2014).
\newblock Concurrent aerodynamic optimization of rotor blades using a nash game
  method.
\newblock \textit{Journal of the American Helicopter Society} .

\bibitem[{Li \& Ba{\c{s}}ar(1987)}]{MR899829}
\textsc{Li, S.} \& \textsc{Ba{\c{s}}ar, T.} (1987).
\newblock Distributed algorithms for the computation of noncooperative
  equilibria.
\newblock \textit{Automatica J. IFAC} \textbf{23}, 523--533.

\bibitem[{Littman \& Stone(2005)}]{littman2005polynomial}
\textsc{Littman, M.~L.} \& \textsc{Stone, P.} (2005).
\newblock A polynomial-time nash equilibrium algorithm for repeated games.
\newblock \textit{Decision Support Systems} \textbf{39}, 55--66.

\bibitem[{McKay et~al.(1979)McKay, Beckman \& Conover}]{mckay1979comparison}
\textsc{McKay, M.~D.}, \textsc{Beckman, R.~J.} \& \textsc{Conover, W.~J.}
  (1979).
\newblock Comparison of three methods for selecting values of input variables
  in the analysis of output from a computer code.
\newblock \textit{Technometrics} \textbf{21}, 239--245.

\bibitem[{Mockus(1989)}]{mockus1989}
\textsc{Mockus, J.} (1989).
\newblock \textit{Bayesian Approach to Global Optimization: Theory and
  Applications}.
\newblock Springer.

\bibitem[{Neyman \& Sorin(2003)}]{neyman2003stochastic}
\textsc{Neyman, A.} \& \textsc{Sorin, S.} (2003).
\newblock \textit{Stochastic games and applications}, vol. 570.
\newblock Springer Science \& Business Media.

\bibitem[{Nishimura et~al.(2009)Nishimura, Hayashi \&
  Fukushima}]{nishimura2009robust}
\textsc{Nishimura, R.}, \textsc{Hayashi, S.} \& \textsc{Fukushima, M.} (2009).
\newblock Robust nash equilibria in n-person non-cooperative games: Uniqueness
  and reformulation.
\newblock \textit{Pacific Journal of Optimization} \textbf{5}, 237--259.

\bibitem[{Parr(2012)}]{Parr2012}
\textsc{Parr, J.~M.} (2012).
\newblock \textit{Improvement Criteria for Constraint Handling and
  Multiobjective Optimization}.
\newblock Ph.D. thesis, University of Southampton.

\bibitem[{Picheny(2014)}]{picheny2014stepwise}
\textsc{Picheny, V.} (2014).
\newblock A stepwise uncertainty reduction approach to constrained global
  optimization.
\newblock In \textit{Proceedings of the 17th International Conference on
  Artificial Intelligence and Statistics}, vol.~33. JMLR W\&CP.

\bibitem[{Picheny \& Binois(2017)}]{picheny2017}
\textsc{Picheny, V.} \& \textsc{Binois, M.} (2017).
\newblock \textit{{GPGame}: Solving Complex Game problems using Gaussian
  processes}.
\newblock R package version 0.1.3.

\bibitem[{Plumlee(2014)}]{Plumlee2014a}
\textsc{Plumlee, M.} (2014).
\newblock Fast prediction of deterministic functions using sparse grid
  experimental designs.
\newblock \textit{Journal of the American Statistical Association}
  \textbf{109}, 1581--1591.

\bibitem[{{R Core Team}(2016)}]{R2016}
\textsc{{R Core Team}} (2016).
\newblock \textit{R: A Language and Environment for Statistical Computing}.
\newblock R Foundation for Statistical Computing, Vienna, Austria.

\bibitem[{Rasmussen \& Williams(2006)}]{Rasmussen2006}
\textsc{Rasmussen, C.~E.} \& \textsc{Williams, C.} (2006).
\newblock \textit{{Gaussian Processes for Machine Learning}}.
\newblock MIT Press.

\bibitem[{Rosenm{\"u}ller(1971)}]{rosenmuller1971generalization}
\textsc{Rosenm{\"u}ller, J.} (1971).
\newblock On a generalization of the lemke--howson algorithm to noncooperative
  n-person games.
\newblock \textit{SIAM Journal on Applied Mathematics} \textbf{21}, 73--79.

\bibitem[{Roustant et~al.(2012)Roustant, Ginsbourger \& Deville}]{Roustant2012}
\textsc{Roustant, O.}, \textsc{Ginsbourger, D.} \& \textsc{Deville, Y.} (2012).
\newblock {DiceKriging}, {DiceOptim}: Two {R} packages for the analysis of
  computer experiments by kriging-based metamodeling and optimization.
\newblock \textit{Journal of Statistical Software} \textbf{51}, 1--55.

\bibitem[{Rulli{\`e}re et~al.(2016)Rulli{\`e}re, Durrande, Bachoc \&
  Chevalier}]{rulliere2016nested}
\textsc{Rulli{\`e}re, D.}, \textsc{Durrande, N.}, \textsc{Bachoc, F.} \&
  \textsc{Chevalier, C.} (2016).
\newblock Nested kriging predictions for datasets with a large number of
  observations.
\newblock \textit{Statistics and Computing} , 1--19.

\bibitem[{{Scilab Enterprises}(2012)}]{Scilab2012}
\textsc{{Scilab Enterprises}} (2012).
\newblock \textit{Scilab: Free and Open Source software for numerical
  computation}.
\newblock Scilab Enterprises, Orsay, France.

\bibitem[{Shahriari et~al.(2016)Shahriari, Swersky, Wang, Adams \&
  de~Freitas}]{Shahriari2016}
\textsc{Shahriari, B.}, \textsc{Swersky, K.}, \textsc{Wang, Z.}, \textsc{Adams,
  R.~P.} \& \textsc{de~Freitas, N.} (2016).
\newblock Taking the human out of the loop: A review of bayesian optimization.
\newblock \textit{Proceedings of the IEEE} \textbf{104}, 148--175.

\bibitem[{Shapley(1953)}]{shapley1953stochastic}
\textsc{Shapley, L.~S.} (1953).
\newblock Stochastic games.
\newblock \textit{Proceedings of the national academy of sciences} \textbf{39},
  1095--1100.

\bibitem[{Srinivas et~al.(2012)Srinivas, Krause, Kakade \&
  Seeger}]{srinivas2012information}
\textsc{Srinivas, N.}, \textsc{Krause, A.}, \textsc{Kakade, S.~M.} \&
  \textsc{Seeger, M.} (2012).
\newblock Information-theoretic regret bounds for gaussian process optimization
  in the bandit setting.
\newblock \textit{Information Theory, IEEE Transactions on} \textbf{58},
  3250--3265.

\bibitem[{Uryas'ev \& Rubinstein(1994)}]{Rubinstein}
\textsc{Uryas'ev, S.} \& \textsc{Rubinstein, R.~Y.} (1994).
\newblock On relaxation algorithms in computation of noncooperative equilibria.
\newblock \textit{IEEE Transactions on Automatic Control} \textbf{39},
  1263--1267.

\bibitem[{Villemonteix et~al.(2009)Villemonteix, Vazquez \&
  Walter}]{villemonteix2009informational}
\textsc{Villemonteix, J.}, \textsc{Vazquez, E.} \& \textsc{Walter, E.} (2009).
\newblock An informational approach to the global optimization of
  expensive-to-evaluate functions.
\newblock \textit{Journal of Global Optimization} \textbf{44}, 509--534.

\bibitem[{Wagner et~al.(2010)Wagner, Emmerich, Deutz \&
  Ponweiser}]{wagner2010expected}
\textsc{Wagner, T.}, \textsc{Emmerich, M.}, \textsc{Deutz, A.} \&
  \textsc{Ponweiser, W.} (2010).
\newblock On expected-improvement criteria for model-based multi-objective
  optimization.
\newblock In \textit{International Conference on Parallel Problem Solving from
  Nature}. Springer.

\bibitem[{Wang \& Shan(2007)}]{Wang2007}
\textsc{Wang, G.} \& \textsc{Shan, S.} (2007).
\newblock Review of metamodeling techniques in support of engineering design
  optimization.
\newblock \textit{Journal of Mechanical Design} \textbf{129}, 370.

\bibitem[{Wilson \& Nickisch(2015)}]{wilson2015kernel}
\textsc{Wilson, A.} \& \textsc{Nickisch, H.} (2015).
\newblock Kernel interpolation for scalable structured gaussian processes
  (kiss-gp).
\newblock In \textit{International Conference on Machine Learning}.

\bibitem[{{\v{Z}}ilinskas \& Zhigljavsky(2016)}]{vzilinskas2016stochastic}
\textsc{{\v{Z}}ilinskas, A.} \& \textsc{Zhigljavsky, A.} (2016).
\newblock Stochastic global optimization: a review on the occasion of 25 years
  of informatica.
\newblock \textit{Informatica} \textbf{27}, 229--256.

\end{thebibliography}

\end{document}